\documentclass[preprint,5p,times,twocolumn,authoryear]{elsarticle}
\usepackage{times}
\usepackage{graphicx}
\usepackage{algorithm}
\usepackage{algorithmic}
\usepackage[cmex10]{amsmath}
\usepackage{amssymb,array}
\usepackage{bm}
\usepackage{subfig}
\usepackage{multirow}
\usepackage{color}

\newcommand{\m}{\mathbf} %define matrix
\newcommand{\pt}[1]{{\color{black}{#1}}}

\journal{Neural Networks}

\begin{document}

\begin{frontmatter}

\title{Probabilistic Learning Vector Quantization on Manifold of Symmetric Positive Definite Matrices}

\author[label1,label2]{Fengzhen Tang \corref{cor1}}
\ead{tangfengzhen@sia.cn}

\author[label1,label2,label3]{Haifeng Feng}
\ead{fenghaifeng@sia.cn}

\author[label4]{Peter Tino}
\ead{P.Tino@cs.bham.ac.uk}

\author[label5]{Bailu Si}
\ead{bailusi@bnu.edu.cn}

\author[label6]{Daxiong Ji}
\ead{jidaxiong@zju.edu.cn}


\cortext[cor1]{Corresponding author}

\address[label1]{State Key Laboratory of Robotics, Shenyang Institute of Automation, Chinese Academy of Sciences, Shenyang, 110016, China}
\address[label2]{Institutes for Robotics and Intelligent Manufacturing, Chinese Academy of Sciences, Shenyang, 110169, China}
\address[label3]{University of Chinese Academy of Sciences, Beijing, 100049, China}

\address[label4]{School of computer Science, University of Birmingham, Birmingham, B15 2TT, UK}

\address[label5]{School of Systems Science, Beijing Normal University, Beijing,100875, China}

\address[label6]{Institute of Marine Electronics and Intelligent Systems, Ocean College, Zhejiang University, The Key Laboratory of Ocean Observation-Imaging Testbed of Zhejiang Province, The Engineering Research Center of Oceanic Sensing Technology and Equipment, Ministry of Education, Zhoushan, 316021, China}

\begin{abstract}
In this paper, we develop a new classification method for manifold-valued data in the framework of probabilistic learning vector quantization.
In many classification scenarios, the data can be naturally represented by symmetric positive definite matrices, which are inherently points that live on a curved Riemannian manifold. Due to the non-Euclidean geometry of Riemannian manifolds, traditional Euclidean machine learning algorithms yield poor results on such data. In this paper, we generalize the probabilistic learning vector quantization algorithm for data points living on the manifold of symmetric positive definite matrices equipped with Riemannian natural metric (affine-invariant metric). By exploiting the induced Riemannian distance, we derive the probabilistic learning Riemannian space quantization algorithm, obtaining the learning rule through Riemannian gradient descent. Empirical investigations on synthetic data, image data , and motor imagery EEG data demonstrate the superior performance of the proposed method. 
\end{abstract}

\begin{keyword}
Probabilistic learning vector quantization \sep learning vector quantization  \sep symmetric positive definite matrices \sep Riemannian geodesic distances   \sep Riemannian manifold 
\end{keyword}

\end{frontmatter}

\section{Introduction}
Symmetric positive definite (SPD) matrices are widely used data structures in many disciplines, e.g. in medical imaging \citep{Pennec2006A} and  computer vision as covariance region descriptors \citep{Tuzel2006Region,Jayasumana2015}, as well as in  brain-computer interface (BCI) \citep{Congedo2017}, etc. 
Endowed with an appropriate metric, SPD matrices form a curved Riemannian manifold. 
%As the manifold of SPD matrices lack a vector space structure, the properties of Euclidean spaces cannot apply. 
Consequently, many popular machine learning algorithms such as linear discriminant analysis (LDA),  learning vector quantization (LVQ), or support vector machines (SVM) cannot be directly applied.   
\pt{One can decide to ignore the nonlinear geometry of the manifold and apply Euclidean machine learning methods directly. However, this approach usually results in poor accuracy \citep{Arsigny2006}. 
A more plausible way of dealing with the nonlinear nature of the SPD manifold relies on first order local approximations of the manifold by  tangent spaces.} For example, \citep{Barachant2012Multiclass} proposed a tangent space linear discriminate analysis (TSLDA) method which projects SPD matrices into the tangent space at their Riemannian geometric mean and then applies standard Euclidean LDA in the tangent space. This idea was further extended in
\citep{Xie2017Motor}, where sub-manifold learning for dimension reduction is used before the tangent space approximation. 
\pt{However, the first-order approximations can lead to undesirable distortion, especially in regions far from the tangent space origin \citep{Tuzel2008,Jayasumana2015}.} The mean of the SPD matrices is a frequently used  candidate for the tangent space origin, however, no theoretical proof exists to guarantee the mean yields the best tangent space approximation for the data \citep{Tuzel2008}. 

\pt{To avoid the above approximation, Gaussian radial basis function kernel with geodesic distance is proposed in \citep{Jayasumana2015}, such that classical kernel methods, e.g.  kernel SVM and kernel LDA, can be used.} However, most of these kernel classifiers are intrinsically binary. The one-vs-all or one-vs-one voting strategy has to be used when multiple classes are involved in the learning problem, making the task very complicated. \pt{Moreover, as the kernel method maps the data into high dimensional Reproducing Kernel Hilbert space (RKHS), implicitly \citep{Burges1998,Jayasumana2015}, the learned classifier is not straightforward to interpret.}

%intoduce learning vector quantization
In Euclidean spaces, 
learning vector quantization (LVQ), first introduced in \citep{Kohonen1986}, has enjoyed great popularity because of its simplicity, intuitive nature, and natural accommodation of multi-class classification problems. It is a prototype and distance based supervised classification algorithm
that has been used in a variety of applications such as image and signal processing, the biomedical field and medicine, and industry \citep{Nova2014A}. 
%The approach is rather intuitive and particularly simple but very powerful. %This approach enjoys a great popularity for several reasons. 
Unlike deep networks, the LVQ system is straightforward to interpret. 
The classifier constructed by LVQ is parametrized by a set of labeled prototypes living in the same space as the training data. The classification of an unknown instance takes place as an inference of the class of the closest prototype in terms of the involved metric. The learning rules of LVQ are typically based on intuitive Hebbian learning, making the implementation of the method very simple.

%Current LVQ and its extensions are designed to deal with data items in the form of finite dimensional real vectors.

% introduce our
\pt{In this paper, we generalize the successful robust soft learning vector quantization method \citep{Seo2003Soft} to the manifold of symmetric positive definite matrices.  The proposed probabilistic learning Riemannian space quantization (PLRSQ) inherits the advantages of LVQ methods, i.e. simple, intuitive, life-long learning, and natural multi-class classifier. The classification method is derived in the probabilistic framework, can produce confidence (probability) for its prediction, and thus can be easily extended to classification with rejection \citep{Fischer2014,Ni2019On} which can be of great use in medical analysis and BCI application. Moreover, the proposed method is implemented by Riemannian stochastic gradient descent algorithm avoiding the problems caused by approximating the data with its projection onto the tangent space at a particular point on the manifold.}

The paper has the following structure:  In Section \ref{sec:relatedwork} we briefly review existing learning algorithms dealing with SPD matrices.  In Section \ref{sec:RSLVQ}, we concisely introduce LVQ and its variant -- probabilistic LVQ. In Section \ref{sec:manifolds} we introduce relevant concepts of the Riemannian manifold of SPD matrices. Our proposed probabilistic learning Riemannian space quantization (PLRSQ) method is described in details in Section \ref{sec:RRSLVQ}. Experimental results are given in Section \ref{sec:exp}. Main findings and conclusions are presented in Section \ref{sec:con}.

\section{Related Work } 
\label{sec:relatedwork}
In this paper, we focus on the Riemannian manifold of SPD matrices. The design of  non-linear method on SPD matrices has strong roots in the field of  diffusion tensor imaging (DTI) \citep{Thomas2004,Pennec2006A,Fletcher2004Principal}. SPD matrices have extensive applications in computer vision \citep{Jayasumana2015}, e.g. covariance region descriptors are widely used in object detection \citep{Tuzel2008}, texture classification \citep{Tuzel2006Region}, face recognition,   and object categorization \citep{Wang2012Covariance}. Recently, SPD matrices become popular in the field of brain-computer interface (BCI) for the design of electroencephalogram (EEG) decoder \citep{Ang2008filter,Congedo2017}. 

The properties and the calculation of the geometric mean of SPD matrices have been studied in  \citep{Moakher2008A}, which subsequently facilitates many learning algorithms for SPD manifold-valued data. Among them, {minimum distance to Riemannian mean (MDRM)} \citep{Barachant2012Multiclass} classification algorithm is one of the most straightforward ones. It is an extension of the minimum distance to mean (MDM) classification algorithm using Riemannian geometric distance and Riemannian geometric mean. MDRM learns a cluster center (i. e. Riemannian geometric mean ) for each class of instances and predicts the class label of an unknown instance by finding the center with shortest Riemnnian geometric distance to the instance. 

Principal geodesic analysis (PGA) presented in  \citep{Thomas2004,Fletcher2004Principal} projects the data into the tangent space at the Riemannian geometric mean of the SPD matrices, and subsequently apply Euclidean principal component analysis (PCA). By the same way of tangent space projection,  {Tangent space linear discriminate analysis (TSLDA)} \citep{Barachant2012Multiclass}, which is an extension of Euclidean linear discriminate analysis to the Riemannian manifold of SPD matrices is developed. In \citep{Xie2017Motor}, tangent space of sub-manifold with linear discriminate analysis or support vector machines is proposed. This method first  learns an optimal map from the original Riemannian space of SPD matrices to the Riemannian sub-manifold through jointly diagonalizing the Riemannian geometric means of the two class data. The optimal map transforms the original SPD matrices into lower dimensional SPD matrices where tangent space projection is performed and then Euclidean LDA or SVM is applied. In \citep{Barachant2013Classification}, a Riemmanian based kernel is constructed using scalar product defined in the tangent plane at a reference point (usually geometric mean) in the manifold.  These aforementioned methods all suffer from the drawback of approximating the manifold by tangent spaces at a reference point in the manifold, and most of the methods are restricted to the binary classification.

In \citep{Jayasumana2015}, log-Euclidean Gaussian kernel on the Riemannian manifold of SPD matrices is proposed, using the log-Euclidean geodesic distance. These kernel methods are free of tangent space approximation, but the computation of kernel matrix is quite heavy (scales quadratically with the number of training examples ), especially when the size of training points is large. Moreover, the learned model of kernel methods is difficult to interpret. 

Our proposed probabilistic learning Riemannian space quantization (PLRSQ) method is different from the existing methods targeted for the SPD matrix-valued data. In particular, our proposed PLRSQ method is implemented by Riemannian stochastic gradient descent algorithm. Unlike TSLDA, it does not need to approximate the data by projections to the tangent space at a particular point on the manifold. Note that the Riemannian stochastic gradient descent algorithm also introduces projections of SPD matrices to the tangent spaces, but only at a local scale. Though the PLRSQ method shares some properties with MDRM, it is potentially far more powerful than MDRM in that, if needed,  PLRSQ can learn multiple representatives for each class, as opposed to only one center per class in MDRM. Even with one prototype per class, the PLRSQ method shows superior performance to MDRM, since MDRM finds the class representatives in an unsupervised manner as class conditional means of the training data. Our method is more computationally efficient than the kernel methods with geodesic distances as the training time of our proposed method scales linearly with the number of training examples. \pt{Moreover, the learned model of our method is easy to understand since explicit ``class representatives" (prototypes) are obtained during the course of training. If desired, interpretability of our method could be enhanced by explicit incorporation of a representability term in the cost function  as in \citep{Hammer2014}.
%, or using  cross-entropy based probabilistic framework as in \citep{2018Probabilistic} PLEASE CHECK, I DON'T QUITE UNDERSTAND HOW CROSS-ENTROPY BASED COST FUNCTION CAN SPECIFICALLY ENHANCE INTERPRETABILITY OF THE PROTOTYPES. BUT I HAVEN'T READ THE PAPER. 
%Our future work will consider extension of these LVQ variants to Riemannian manifolds.
}

%do not approximate the data with its projection to a tangent space at a particular point on the manifold like methods such TSLDA, even though we do need to calculate the projection of the SPD matrix to its tangent space. The projection is introduced by Riemannian stochastic gradient descent algorithm. It does not cause the distortion problem, as our method only manipulates a small neighborhood of the projection origin. Besides, the choice of the projection origin is not a problem in our method, as the project origin will be the prototype to be updated. Different from kernel methods, our method is more computationally efficient, training time scaling linearly with the number of training examples. Moreover, the learned model of our method is interpretable. 

\section{\pt{Robust Soft Leaning Vector Quantization}}
\label{sec:RSLVQ}
Our approach is developed within the framework of learning vector quantization (LVQ) \citep{Kohonen1986}. In this section, we will briefly introduce LVQ.

Consider a training dataset $(\bm{x}_i,y_i) \in \mathbb{R}^n \times \{1,...,C\}$, $i=1,..,m$, where $n$ is the dimension of the inputs, $C$ is the number of different classes and $m$ is the number of training examples. A typical LVQ classifier consists $M$ ($M\geq C$) prototypes $\bm{w}_i \in \mathbb{R}^n$, which are labeled by $c(\bm{w}_i) \in \{1,...,C\}$.  The set of labeled prototype vectors is denoted as $\mathcal{T} = \{(\bm{w}_i,c_i)\}_{i=1}^M$ in this paper.
The classification is implemented as a winner-takes-all scheme. For a data point $\bm{x} \in \mathbb{R}^n$, the output class is determined by the class label of its closest prototype: i.e. $\hat y(\bm{x}) := c(\bm{w}_i)$ such that %$ \bm{w}_i = \arg \min_{\bm{w}_j} d(\bm{x},\bm{w}_j)$, 
$ i = \arg \min_{j} d(\bm{x},\bm{w}_j)$,
where $d(\cdot, \cdot)$ is a distance measure in $\mathbb{R}^n$. 
%{Each labeled prototype $\bm{w}_i$ with label $c(\bm{w}_i)$ defines a receptive filed in the input space -- a set of points which pick this prototype as their winner. The goal of learning the LVQ classifier is to adapt prototypes automatically such that the class labels of data points within the receptive field coincide with the label of the respective prototype as much as possible.}

%\begin{figure}
%\centering
%\includegraphics[scale=0.3]{LVQ.eps}
%\caption{ Illustration of LVQ. Assume there are 3 classes, each class there is one prototype assigned to it. }
%\label{fig:LVQ}
%\end{figure}

There are many variants of LVQ algorithm. A detailed review of LVQ method was given in \citep{Nova2014A}. Here we will describe the robust soft  learning vector quantization (RSLVQ) based on likelihood ratio maximization \citep{Seo2003Soft,Seo2003Softa}.  An alternative generalization of LVQ was termed generalized learning vector quantization (GLVQ), which was based on margin maximization \citep{Sato1996}. \pt{Compared to GLVQ, RSLVQ derived in a probabilistic approach, is more flexible in the case of overlapping classes \citep{Nova2014A}.}

RSLVQ algorithm learns the prototype locations based on a statistic modeling of the given data distribution, i. e. the probability density of the data is described by a mixture model. It is assume that each component $j$ of the mixture generates data belonging to and only to one of the $C$ classes denoted as $c_j$. The probability density $p(\bm{x})$ of the data points $\bm{x}$ is modeled as follows:

\begin{eqnarray}
p(\bm{x}|\mathcal{T}) = \sum_{y=1}^{C} \sum_{\{j:c_j = y\}}
 p(\bm{x}|j)P(j)
\end{eqnarray}

\noindent
Here, $P(j)$ is the probability that data points are generated by a particular component $j$ of the mixture. % and it can be chosen identically for each component.
$p(\bm{x}|j)$ is the conditional probability that this component $j$ generates a particular data point $\bm{x}$. The conditional probability $p(\bm{x}|j)$ is a function of prototypes $\bm{w}_j$, which is usually interpreted as the representative feature vector for all data points generated by component $j$. A possible choice for $p(\bm{x}|j)$ is the normalized exponential form $p(\bm{x}|j) = K(j) \exp f(\bm{x}, \bm{w}_j)$. In \citep{Seo2003Soft}, a Gaussian mixture is assumed, i. e. $K(j) = (2\pi \sigma_j^2)^{-n/2}$ and $f(\bm{x}, \bm{w}_j) = - d(\bm{x}, \m{w}_j)/2\sigma_j^2$,  where $d(\cdot, \cdot)$ is the squared Euclidean distance,  and  every component is assumed to have equal variance $\sigma_j^2 = \sigma^2$ and equal prior probability $P(j) = 1/m$ for all $j$.
RSLVQ learns prototypes by maximizing the likelihood ratio
$$
L =\prod_{i=1}^m \frac{p(\bm{x}_i,y_i|\mathcal{T})}{p(\bm{x}_i|\mathcal{T})}
$$
via stochastic gradient ascent algorithm. $p(\bm{x},y|\mathcal{T}) = \sum_{\{j:c_j = y\}} p(\bm{x}|j)P(j)$ is the probability density that a data point $\bm{x}$ is generated by the mixture model for the correct class. 
The learning rule of prototypes in the presence of one example $(\bm{x},y)$ is obtained by computing the derivative of the log likelihood ratio with respect to $\bm{w}_j$ (see \citep{Seo2003Soft}):

 \begin{eqnarray}
\Delta \bm{w}_j  =  \frac{\alpha}{\sigma^2}  \left\{ 
             \begin{array}{lcl}
             {(P_y(j|\bm{x}) - P(j|x))} (\bm{x} - \bm{w}_j) &\text{if} &  c_j = y \\
             {-P(j|\bm{x})} (\bm{x} - \bm{w}_j) &\text{if} &c_j \neq y 
             \end{array}  
        \right.
\end{eqnarray}

\noindent
where $0<\alpha<1$ is the learning rate, $P_y(l|\bm{x})$  and $ P(l|\bm{x})$ are assignment probabilities
  \begin{eqnarray}
 P_y(l|\bm{x}) &=& \frac{\exp f(\bm{x},\bm{w}_j)}{\sum_{\{j:c_j=y \}}\exp f(\bm{x},\bm{w}_j)}
 \end{eqnarray}
 
  \begin{eqnarray}
 P(l|\bm{x}) &=& \frac{ \exp f(\bm{x},\bm{w}_j)}{\sum_{j=1}^M \exp f(\bm{x},\bm{w}_j)}
 \end{eqnarray}
The learning rule reflects the fact that prototypes with the same label as that of  the data point are attracted to the data point, while prototypes with different labels from the data point are repelled. 
% 
%representing the posterior probability that the data point $\bm{x}$ is assigned to the component $l$ of the mixture, given that the data point was generated by the correct class, and $P(l|x)$ is also assignment probability:
%

% 
%describing the posterior probability that the data point $\bm{x}$ is assigned to the component $l$ of the complete mixture using all classes. Note that $(P_y(l|\bm{x}) - P(l|x)) $ is always positive. 

\pt{
We note that the log-likelihood ratio used in
RSLVQ can be extended to a more natural cross-entropy cost function employed in Probabilistic LVQ (PLVQ) \citep{2018Probabilistic}. In fact, PLVQ coincides with RSLVQ if the only stochastic component in the joint distribution $p(\bm{x},y)$ over $\mathbb{R}^{n} \times \{ 1,2,...,C \}$ is the marginal over the inputs $p(\bm{x})$
and the input-conditional class distributions $p(y|\bm{x})$ are delta-functions (see \citep{2018Probabilistic}).
The framework of PLVQ is preferable in cases of genuine class uncertainty in (at least some regions of) the input space, leading to more representative class prototypes.
Even though we derive our method as an extension of RSLVQ to Riemannian manifolds, the core ideas of our method can be applied to PLVQ as well.  
}

The above described RSLVQ is designed for classification of vector-valued data, with the assumption of Gaussian mixture model which relies on the Euclidean distance between the input pattern $\bm{x}$ and the prototype $\bm{w}$, i. e. $f(\bm{x},\bm{w}) = \frac{-\parallel \bm{x} - \bm{w} \parallel^2}{2\sigma^2}$. In the following sections, we generalize this method to deal with data points that live in the Riemannian manifold of symmetric positive definite matrices, where the function  $f(\cdot,\cdot)$ depends on the Riemannian geodesic distance between the input pattern and the prototype.

\section{Riemannian Manifold of SPD Matrices}
\label{sec:manifolds}
Each $n \times n$ real \emph{symmetric positive definite (SPD) matrix} has the property: $\bm{v}^T \m{X} \bm{v} \geq 0$ for all nonzero $\bm{v} \in \mathbb{R}^n$. The space $\mathbb{S}^{+}(n)$ of all $n \times n$ SPD matrices is not a vector space, since an SPD matrix when multiplied by a negative scalar is no longer SPD. In fact, $\mathbb{S}^{+}(n)$ forms a convex cone in the $n^2$-dimensional Euclidean space. Hence, Euclidean metric is no longer suitable to describe its geometry. A Riemannian metric can be introduced to $\mathbb{S}^{+}(n)$, making $\mathbb{S}^{+}(n)$ a curved Riemannian manifold \citep{Thomas2004,Pennec2006A,Arsigny2006,Jayasumana2015}. 
%Riemannian metrics provide better description on the geometry of the space   $\mathbb{S}^{+}(n)$, as they introduce an infinite distance between an SPD matrix and non-SPD matrix \citep{Pennec2006A,Arsigny2006,Jayasumana2015}. 
Two popular Riemannian metrics have been proposed on $\mathbb{S}^{+}(n)$, namely affine-invariant Riemannian metric \citep{Pennec2006A} and Log-Euclidean metric \citep{Arsigny2006}. The affine-invariant Riemannian metric is also called the Riemannian natural metric and is the main focus of this paper. 

Let  $T_\m{X} \mathbb{S}^{+}(n)$ denote the tangent space to $\mathbb{S}^{+}(n)$ at point $\m{X} \in  \mathbb{S}^{+}(n)$. 
%Tangent space  $T_\m{X} \mathbb{S}^{+}(n)$ is a vector space containing tangent vectors of all possible curves on the manifold passing through that point $\m{X}$. 
The affine-invariant Riemannian metric or Riemannian natural metric is defined as follows: 
 \begin{equation}
 \label{eq:RieNatnmetric}
 \langle \m{V}_1,\m{V}_2\rangle_\m{X} = \text{Tr} (\m{V}_1 \m{X}^{-1}\m{V}_2 \m{X}^{-1})
 \end{equation} 
where $\m{V}_1$, $\m{V}_2 \in T_\m{X} \mathbb{S}^{+}(n)$ and $\text{Tr}$ is the  trace operator.
 Note that the tangent space $T_\m{X} \mathbb{S}^{+}(n)$ is the space $\mathbb{S}(n)$ of symmetric matrices.
With the introduction of the Riemannian metric, many geometric notions can be defined,  for instance, the length of a curve on the manifold.
%, which is defined as the integral of the lengths of all tangent vectors to the curve. 
Formally, a \emph{curve} on $\mathbb{S}^{+}(n)$  is a differentiable path connecting two points $\m{X}_1, \m{X}_2 \in \mathbb{S}^{+}(n)$, i. e. $\gamma (t) : [0,1] \rightarrow \mathbb{S}^{+}(n)$ with $ \gamma (0) = \m{X}_1$, $ \gamma (1) = \m{X}_2$, and $\dot{\gamma}(t) \in T_{ \gamma (t)} \mathbb{S}^{+}(n)$.
The length of the curve is defined as follows:
\begin{equation}
\label{eq:curvLen}
L(\gamma(t)) = \int_0^1 \parallel \dot{\gamma} (t) \parallel_{\gamma(t)} dt
%L(\gamma(t)) = \int_0^1  \sqrt{\langle\dot{\gamma} (t), \dot{\gamma} (t)\rangle_{\gamma(t)}} dt
%L(\phi(t)) = \int_0^1  \sqrt{\langle\dot{\phi} (t), \dot{\phi} (t)\rangle_{\phi(t)}} dt
\end{equation}
\noindent
where $\parallel \cdot \parallel_{\m{X}}$ is the norm induced by the inner product $\langle \cdot, \cdot\rangle_{\m{X}}$.
%For any tangent vector $\m{V} \in T_{\m{X}} \mathbb{S}^{+}(n)$, it has length  $\parallel \m{V}  \parallel_{\m{X}} = \sqrt{\langle\m{V}, \m{V}\rangle_{\m{X}}}$.

A naturally parameterized curve (i.e. parametrized by arc length) that minimizes the distance between two points on the manifold is called \emph{geodesic curve}.  A geodesic curve at the point $\m{X}$ in the direction of $ \m{V} \in T_\m{X} \mathbb{S}^+(n)$ on  $\mathbb{S}^{+}(n)$ has an analytic expression :
\begin{equation}
%\gamma_{\m{X},\m{V}} (t) =  \m{X}^{1/2} \exp (t \m{X}^{-1/2} \m{V}\m{X}^{-1/2})\m{X}^{1/2}
\gamma (t) =  \m{X}^{1/2}  \exp (t \m{X}^{-1/2} \m{V}\m{X}^{-1/2})\m{X}^{1/2}
\label{eq:geoCurve}
\end{equation}
\noindent
where $\exp$ is the exponential of matrix. For a symmetric matrix $\m{V} \in \mathbb{S}(n)$, the matrix exponential $\exp(\m{V})$ can be computed via eigenvalue decomposition:
$
\exp(\m{V}) =  {\bf U} \ \text{diag}(\exp(\lambda_1),...,\exp(\lambda_n)) \ {\bf U}^T,
$ 
where $\lambda_1,...,\lambda_n$ are eigenvalues of $\m{V}$,  $\bf U$ is the matrix of eigenvectors of ${\bf V}$.
%, and $\text{diag}$ denotes the diagonal matrix with the arguments of $\text{diag}$ being the diagonal elements of the matrix. 

%Obviously, this geodesic curve is entirely contained in the manifold. For any given pair $\m{X}_1$,  $\m{X}_2 \in \mathbb{S}^+(n)$, we can find a geodesic curve $\gamma (t)$  , such that $\gamma(0) = \m{X}_1$ and $\gamma(1)= \m{X}_2$, by taking the initial velocity $\dot{\gamma} (0) = \m{X}^{1/2} \text{Log} (\m{X}^{-1/2} \m{X}_2\m{X}^{-1/2})\m{X}^{1/2} \in  T_\m{X} \mathbb{S}^+(n)$ \citep{Duan2014Riemannian}. 

The \emph{geodesic distance} between two points on the manifold is the length of the geodesic curve connecting them. On $\mathbb{S}^+(n)$, it reads \citep{Moakher2008A}:
\begin{equation}
\label{eq:Riemaniandistance}
\delta(\m{X}_1, \m{X}_2)  = \parallel  \log (\m{X}_1^{-1} \m{X}_2)\parallel_F = \left[\sum_{i = 1}^n \log^2 \lambda_i \right]^{1/2}
\end{equation}
where  $\log$ denotes the principal logarithm of matrix, $\parallel \cdot \parallel_F$ represents the Frobenius norm,  and  $\lambda_i, i=1,...,n$ are the real eigenvalues of $\m{X}_1^{-1} \m{X}_2$.  Analogously to $\exp$, we have  $\log({\bf X}) =  {\bf U} \ \text{diag}(\log(\lambda_1),...,\log(\lambda_n)) \ {\bf U}^T$.
An important property of this geodesic distance is that it is affine-invariant, i.e.  $\delta( \m{W}^T \m{X}_1 \m{W}, \m{W}^T\m{X}_2 \m{W} ) = \delta(\m{X}_1, \m{X}_2), \forall \ \m{W} \in Gl(n)$ where $Gl(n)$ represents the general linear group, consisting of all nonsingular real matrices of rank $n$.
% It implies classifiers based on this distance are robust to linear transformations. 

{Denote the \emph{Riemannian gradient} of a smooth real-valued function $f: \mathbb{S}^+(n) \mapsto \mathbb{R}$ at a point $\m{X} \in \mathbb{S}^+(n)$ by $\nabla _{\m{X}} f$. 
Given a smooth curve $\gamma: \mathbb{R} \to \mathbb{S}^+(n)$ on $\mathbb{S}^+(n)$,  the composite function $f \circ \gamma: t \mapsto f(\gamma(t)) $ is a smooth function from $\mathbb{R}$ to $\mathbb{R}$ with a well-defined classical derivative. 
The Riemannian gradient $\nabla _{\m{X}} f$ is the unique tangent vector in $T_{\m{X}} \mathbb{S}^+(n)$ satisfying
\begin{equation}
\label{eq:Rgradient}
\langle \dot{\gamma}(0), \nabla_{\m{X}} f\rangle_{\m{X}} = \frac{\text{d}}{\text{d}t} f(\gamma (t))|_{t=0},
\end{equation}
for all curves $\gamma$ such that ${\gamma}(0) = \m{X}$.
Thus, the computation of Riemannian gradient can be performed through the calculation of the classical derivative of the composite function $f \circ \gamma$. }

%The \emph{Riemannian gradient} of a smooth real-valued function $f: \mathbb{S}^+(n) \mapsto \mathbb{R}$ at point $\m{X} \in \mathbb{S}^+(n)$ in the direction of the vector $\m{V}\in T_{\m{X}} \mathbb{S}^+(n)$ is denoted by $\nabla _{\m{X}} f$. Loosely speaking, it quantifies the rate of change of $f$ at $\m{X}$ in the direction $\m{V}$. Given a curve $\gamma$ on $\mathbb{S}^+(n)$,  the composite function $f \circ \gamma: t \mapsto f(\gamma(t)) $ is a smooth function from $\mathbb{R}$ to $\mathbb{R}$ with a well-defined classical derivative. Assuming  ${\gamma}(0) = \m{X}$  and $\dot{\gamma}(0) = \m{V}$, the Riemannian gradient $\nabla _{\m{X}} f$ is the unique tangent vector in $T_{\m{X}} \mathbb{S}^+(n)$ such that
%\begin{equation}
%\label{eq:Rgradient}
%\langle\m{V}, \nabla_{\m{X}} f\rangle_{\m{X}} = \frac{\text{d}}{\text{d}t} f(\gamma (t))|_{t=0}.
%\end{equation}

In a sufficiently small neighborhood $B$ of $\m{X}$ in $\mathbb{S}^+(n)$, it is possible for each point $\m{X}_i \in B$ to identify the tangent vector $\m{V}_i \in   {T}_{\m{X}} \mathbb{S}^+(n)$, such that $\m{V}_i = \dot{\gamma}(0)$ and $\gamma (t)$ the geodesic curve between $\m{X}$ and $\m{X}_i$. The \emph{Riemannian logarithm map} operator $\text{Log}_{\bf X} :  B \rightarrow  T_{\m{X}} \mathbb{S}^+(n)$ maps B on the manifold to the tangent space at $\m{X}$,
i. e.  $\text{Log}_\m{X}(\m{X}_i) = \m{V}_i$. The \emph{Riemannian exponential map} $\text{Exp}_\m{X}:  T_{\m{X}} \mathbb{S}^+(n)  \rightarrow  B$ is the inverse of Log, $\text{Exp}_\m{X}(\m{V}_i) = \gamma(1) = \m{X}_i$. In particular,
\begin{eqnarray}
\text{Exp}_\m{X} (\m{V}_i) = \m{X}^{1/2} \ \exp  (\m{X}^{-1/2} \m{V}_i \m{X}^{-1/2}) \ \m{X}^{1/2} \label{eq:EXPmapSPD}\\
\text{Log}_\m{X} (\m{X}_i) = \m{X}^{1/2} \log  (\m{X}^{-1/2} \m{X}_i \m{X}^{-1/2}) \m{X}^{1/2}.
 \label{eq:LogmapSPD}
\end{eqnarray}

\noindent
The logarithm map provides a way to obtain the tangent vector given two points in the manifold while the exponential map provides a way to access to the corresponding point on the manifold, given a tangent vector. With the definition of Riemannian exponential map and logarithm map, we can transit between the manifold and the tangent space and perform Riemannian gradient descent algorithms on the manifold.

\section{Probabilistic Learning Vector Quantization on the Riemannian Manifold of SPD Matrices }
\label{sec:RRSLVQ}
In this section we present a generalization of the probabilistic learning vector quantization  to the Riemannian space of SPD matrices. 
Consider a $C$-class labeled data set $\{(\m{X}_i,y_i)\}_{i=1}^m$, $\m{X}_i \in \mathbb{S}^+(n)$, $y_i \in \{1,...,C\}$. The classifier consists of a set of $M <m$  labeled prototypes living on 
$\mathbb{S}^+(n)$
, denoted by $\mathcal{W}  = \{ (\m{W}_j, c_j) \}_{j=1}^M$.

Since the concept of Gaussian distributions and mixture of Gaussian distributions can be generalized to the manifold of symmetric positive-definite matrices \citep{Said2017},  following \citep{Seo2003Soft}, we assume that the marginal probability density $p(\m{X}) $ on $\mathbb{S}^+(n)$ that generated the data 
can be approximated by a Gaussian-like mixture model, with one component 
$p(\m{X}|j)$ per prototype $\m{W}_j$ (there can be several prototypes per class):
\begin{eqnarray}
p(\m{X}|\mathcal{W}) = \sum_{y=1}^{C} \sum_{\{j:c_j = y\}}
 p(\m{X}|j) \ P(j).
\end{eqnarray}
\noindent
where the conditional probability $p(\m{X}|j)$ is a Gaussian-like function of the prototype $\m{W}_j$,
\begin{equation}
p(\m{X}|j) \propto e^{f(\m{X}, \m{W}_j) }
\end{equation}

\noindent
with $f(\m{X}, \m{W}_j) = \frac{  - \delta^2(\m{X}, \m{W}_j)}{2\sigma^2} $ and $\sigma^2>0$ being a scale constant. 

Let us consider a data point $\m{X}$ and its true class label $y$. The probability density that a data point $\m{X}$ is generated by the mixture model for the correct class, i.e. the class denoted by $y$ is given as follows:

\begin{eqnarray}
p(\m{X},y|\mathcal{W}) = \sum_{\{j:c_j = y\}}
 p(\m{X}|j)P(j)
\end{eqnarray}

\noindent
Following \citep{Seo2003Soft}, we can maximize the likelihood ratio to obtain the learning rule of prototypes: 

\begin{eqnarray}
\label{eq:likihhodRatio}
L_r =  \prod_{i=1}^m \frac{p(\m{X}_i,y|\mathcal{W})}{p(\m{X}_i|\mathcal{W})}
\end{eqnarray}

\noindent Note that, according to Bayes' theorem:
\begin{equation}
p(y|\m{X};\mathcal{W}) = \frac{p(\m{X},y|\mathcal{W})}{p(\m{X}|\mathcal{W})} = \frac{\sum_{\{j:c_j = y\}}  P(j)e^{f(\m{X}, \m{W}_j) }}{\sum_{i = 1}^M  P(i) e^{f(\m{X}, \m{W}_i)}}
\label{eq:pycon_X}
\end{equation}
where $p(y|\m{X};\mathcal{W})$ represents the conditional probability of assigning an label $y$ to  the input $ \m{X}$. Thus the likelihood ratio given by Eq.~\eqref{eq:likihhodRatio} is the same as the likelihood  function:
\begin{equation}
L = \prod_{i=1}^m p(y_i|\m{X}_i; \mathcal{W})
\end{equation}

\noindent
The learning rule of the prototypes can be obtained by maximization of the log likelihood, which is equivalent to minimization of the negative log likelihood, via  stochastic Riemannian gradient descent \citep{Bonnabel2013Stochastic}. The objective function, i. e.  the negative log likelihood, is given as follows:

\begin{eqnarray}
\label{eq:cost}
E &=& - \log L \nonumber \\
&=& - \sum_{i=1}^m \log   \frac{\sum_{\{j:c_j = y_i\}} P(j) e^{ f(\m{X}_i, \m{W}_j) }}{\sum_{j = 1}^M P(j) e^{f(\m{X}_i, \m{W}_j)}} \nonumber \\
 &=&   \sum_{i=1}^m \left\{  - \log \sum_{\{j:c_j = y_i\}}  P(j) e^{f(\m{X}_i, \m{W}_j) } \right. \nonumber \\
 && \left.+ \log \sum_{j = 1}^M P(j) e^{f(\m{X}_i, \m{W}_j)} \right\} 
\end{eqnarray}

%Consider a data point $\m{X}$ and a class label $y$. We have,
%\begin{eqnarray}
%p(\m{X},y|\mathcal{W}) = \sum_{\{j:c_j = y\}}
% p(\m{X}|j)p(j)
%\end{eqnarray}
%and following
%\noindent
%\cite{Seo2003Soft}, we maximize the likelihood 
%%\begin{equation}
%$
%L = \prod_{i=1}^m p(y_i|\m{X}_i; \mathcal{W}),
%$
%%\end{equation}
%where
%$$
%p(y|\m{X};\mathcal{W}) = \frac{p(\m{X},y|\mathcal{W})}{p(\m{X}|\mathcal{W})} = \frac{\sum_{\{j:c_j = y\}}  p(j)e^{f(\m{X}, \m{W}_j) }}{\sum_{i = 1}^M  p(i) e^{f(\m{X}, \m{W}_i)}}.
%$$
%
%\noindent
%The learning rule of the prototypes can be obtained through minimization of the negative log likelihood, $E =- \log L$ via  stochastic Riemannian gradient descent \cite{Bonnabel2013Stochastic}.
%
%\begin{eqnarray}
%\label{eq:cost}
%E &=&  - \sum_{i=1}^m \log   \frac{\sum_{\{j:c_j = y_i\}} p(j) e^{ f(\m{X}_i, \m{W}_j) }}{\sum_{j = 1}^M p(j) e^{f(\m{X}_i, \m{W}_j)}} \nonumber \\
% &=&   \sum_{i=1}^m \left\{  - \log \sum_{\{j:c_j = y_i\}}  p(j) e^{f(\m{X}_i, \m{W}_j) } \right. \nonumber \\
% && \left.+ \log \sum_{j = 1}^M p(j) e^{f(\m{X}_i, \m{W}_j)} \right\} \nonumber
%\end{eqnarray}

According to Eq.~\eqref{eq:Rgradient}, the Riemannian gradient of the objective function can be computed by the classical derivative of the objective function along the geodesic curve on the manifold. Given an example $\m{X}$ with label $y$, let $\gamma_l(t)$ be a geodesic curve emitting from the $l$-th prototype $\m{W}_l $, 
$l \in \{1,...,M\}$, in the direction of $\m{V}_l \in T_{\m{W}_l} \mathbb{S}^+(n)$, according to the definition of geodesic curves given by Eq.~\eqref{eq:geoCurve}, 
$$\gamma_l (t) =  \m{W}_l^{1/2}  \exp (t \m{W}_l^{-1/2} \m{V}_l\m{W}_l^{-1/2})\m{W}_l^{1/2}.$$

\noindent
If $c_l = y$, the objective function for this example along the curve $\gamma_l(t)$ reads: 
\begin{eqnarray*}
&&\xi(\m{W}_1,...,\gamma_l(t),...,\m{W}_M) \nonumber \\ 
&=&  -   \log \left( \sum_{\{j:c_j = y, j\neq l\}}  P(j) e^{f(\m{X}, \m{W}_j) } +  P(l) e^{f(\m{X},\gamma_l(t))} \right) \nonumber \\  
&& + \log \left( \sum_{j = 1,j\neq l}^M P(j)e^{f({\bf X}, \m{W}_j)}  + P(l) e^{f(\m{X},\gamma_l(t))} \right)
\end{eqnarray*}

\noindent
Then, we can compute the Riemannian gradient of the objective function $\xi$ at point $\m{W}_l$ denoted as $\nabla_{\m{W}_l} \xi $ as follows: 

\begin{eqnarray*}
&& \langle \m{V}_l, \nabla_{\m{W}_l} \xi \rangle_{\m{W}_l} = \frac{ \text{d}}{\text{d}t} \xi (\m{W}_1,...,\gamma_l(t),...,\m{W}_M) \bigg|_{t=0}   \nonumber \\
&=&  - \frac{P(l)e^{f(\m{X}, \gamma_l(t))} (-\frac{1}{2\sigma^2}) \frac{\text{d} { \delta^2 \left(\m{X}, \gamma_l(t)\right) } }{\text{d} t} }{\sum_{\{j:c_j = y,j\neq l \}} P(j) e^{f(\m{X}, \m{W}_j)} +P(l) e^{f(\m{X},\gamma_l(t))} } \bigg|_{t=0}    \nonumber\\ 
&&  +\frac{P(l) e^{f(\m{X}, \gamma_l(t))} (-\frac{1}{2\sigma^2})  \frac{\text{d} { \delta^2 \left(\m{X}, \gamma_l(t)\right) } }{\text{d} t} }{\sum_{j=1,j\neq l}^M  P(j) e^{f(\m{X}, \m{W}_j) }+ P(l) e^{f(\m{X},\gamma_l(t))}}  \bigg|_{t=0} 
\end{eqnarray*}

\noindent 
Denote by $P(l|\m{X},y)$ and $P(l|\m{X})$ 
the posterior probabilities that the data point $\m{X}$ is assigned to the component $l$ of the mixture among the prototypes belonging to class $y$ and among all prototypes, respectively:

$$P(l|\m{X},y) = \frac{P(l) e^{f(\m{X}, \m{W}_l)}}{\sum_{\{j:c_j = y\}}  P(j)e^{f(\m{X}, \m{W}_j) }}$$ 

$$P(l|\m{X}) = \frac{P(l)e^{f(\m{X}, \m{W}_l)}}{\sum_{j=1}^M  P(j)e^{f(\m{X}, \m{W}_j) }}$$

\noindent
Then we can simplify the calculation of $\langle \m{V}_l, \nabla_{\m{W}_l} \xi \rangle_{\m{W}_l}$ as follows:
 
\begin{eqnarray}
&& \langle \m{V}_l, \nabla_{\m{W}_l} \xi \rangle_{\m{W}_l} \nonumber\\
&=&  \frac{1}{2\sigma^2} ( P(l|\m{X},y) - P(l|\m{X}))  \frac{\text{d} { \delta^2\left(\m{X}, \gamma_l(t)\right) } }{\text{d} t} \bigg|_{t=0} 
\end{eqnarray}

Note that $P(l|\m{X},y)$ describes the posterior probability that the data point $\m{X}$ is assigned to the component $l$ of the mixture, given that the data point was generated by the correct class $y$. According to Bayes' rule, $P(l | \m{X},y) = P(l,y | \m{X})/P(y|\m{X})$, where $P(y|\m{X})$ is given in \eqref{eq:pycon_X}. Since the prototype $\m{W}_l$ here also belongs to class $y$, if $\m{W}_l$ is picked, class $y$ will be automatically picked. Therefore we have $P(l,y | \m{X}) = P(l|\m{X})$. Consequently, we can get above expression for $ P(l|\m{X},y)$.

%$P(l|\m{X})$ denotes the posterior of mixture component $l$ in the mixture
%
%, since  If $\m{W}_l$ is picked, class $y$ will be automatically picked.   
%
% which represents the posterior of mixture component $l$ in the mixture, since we know that $W_l$ is prototype of class y. So if you pick $W_l$, you automatically also pick class y.
%
%$P(l|X) = P(l) exp{f(X,W_l)}/p(X)$
%
% Putting things together you will get expression for $ P_y(l|X)$.

If $c_l \neq y$, the objective function  is given as follows: 
\begin{eqnarray*}
&&\xi(\m{W}_1,...,\gamma_l(t),...,\m{W}_M) \nonumber \\ 
&=&  -   \log  \sum_{\{j:c_j = y\}}  P(j) e^{f(\m{X}, \m{W}_j) }  \nonumber \\  
&& + \log \left( \sum_{j = 1,j\neq l}^M P(j)e^{f({\bf X}, \m{W}_j)}  + P(l) e^{f(\m{X},\gamma_l(t))} \right) \label{eq:tderiRie}
\end{eqnarray*}

\noindent
and  Riemannian gradient $\nabla_{\m{W}_l} \xi $ is computed as follows:
\begin{eqnarray}
&& \langle \m{V}_l, \nabla_{\m{W}_l} \xi \rangle_{\m{W}_l} = \frac{ \text{d}}{\text{d}t} \xi (\m{W}_1,...,\gamma_l(t),...,\m{W}_M) \bigg|_{t=0}   \nonumber \\
&=& \frac{P(l) e^{f(\m{X}, \gamma_l(t))} (-\frac{1}{2\sigma^2})  \frac{\text{d} { \delta^2 \left(\m{X}, \gamma_l(t)\right) } }{\text{d} t} }{\sum_{j=1,j\neq l}^M P(j) e^{f(\m{X}, \m{W}_j) }+ P(l) e^{f(\m{X},\gamma_l(t))}}  \bigg|_{t=0} \nonumber\\
&=& - \frac{1}{2\sigma^2}  P(l|\m{X})  \frac{\text{d} {\delta^2\left(\m{X}, \gamma_l(t)\right) } }{\text{d} t} \bigg|_{t=0}
 \label{eq:tderiRieny}
\end{eqnarray}

Denote the Riemannian gradient of the squared Riemannian distance at the point $\m{W}_l$ by $\nabla_{\m{W}_l} \delta_l^2$. We  have:
\begin{equation}
\label{eq:distanceGradient}
\langle\m{V}_l, \nabla_{\m{W}_l} \delta_l^2 \rangle_{\m{W}_l} = \frac{\text{d} {\delta^2\left(\m{X}, \gamma_l(t)\right) } }{\text{d} t} \bigg|_{t=0} 
\end{equation}

\noindent
The Riemannian gradients of the objective function  $\nabla_{\m{W}_l}\xi$  are then

\begin{equation}
\label{eq:RgradentWl}
\nabla_{\m{W}_l} \xi = \frac{\nabla_{\m{W}_l} \delta_l^2}{2\sigma^2} \left\{ \begin{array}{lcc}
   \left(P(l|\m{X},y) - P(l|\m{X})\right), &\text{if}& c_l = y \\ 
 -   P(l|\m{X}) , &\text{if}& c_l \neq y
\end{array}\right.
\end{equation}

\noindent
The Riemannian gradient $\nabla_{\m{W}_l} \delta_l^2$ is given as follows: 

\begin{equation}
\nabla_{\m{W}_l} \delta_l^2 = - 2 \text{Log}_{\m{W}_l} (\m{X})  \label{eq:tderiRieWJF}
\end{equation}

\noindent
with $\text{Log}_{\m{W}_l} ( \cdot)$ defined by Eq. \eqref{eq:LogmapSPD}.
Detailed calculation of the Riemannian Gradient  $\nabla_{\m{W}_l} \delta_l^2$ is given in the \ref{app:gradientcompute}. 
\noindent
Substituting \eqref{eq:tderiRieWJF} into \eqref{eq:RgradentWl}, we can obtain:

\begin{eqnarray*}
&&\nabla_{\m{W}_l} \xi = \frac{1}{\sigma^2} \cdot\\
&&  \left\{ \begin{array}{lc}
-  \left(P(l|\m{X},y) - P(l|\m{X})\right)\text{Log}_{\m{W}_l} (\m{X}), & c_l = y \\ 
   P(l|\m{X}) \text{Log}_{\m{W}_l} (\m{X}), & c_l \neq y
\end{array}\right.
\end{eqnarray*} 

\noindent
and through moving along  Riemannian gradient in the tangent space and mapping back onto the manifold (through Riemannian exponential map), we obtain the prototype learning rule:
%
%\begin{eqnarray*}
%&&\nabla_{\m{W}_l} \xi = \frac{1}{\sigma^2} \cdot\\
%&&  \left\{ \begin{array}{lc}
%-  \left(P_y(l|\m{X}) - P(l|\m{X})\right)\text{Log}_{\m{W}_l} (\m{X}), & c_l = y \\ 
%   P(l|\m{X}) \text{Log}_{\m{W}_l} (\m{X}), & c_l \neq y
%\end{array}\right.
%\end{eqnarray*} 
%
 \begin{eqnarray}
\label{eq:plvqupdatruespd}
&&\m{W}_l^{new} =  \text{Exp}_{\m{W}_l}\left[ \frac{\alpha }{\sigma^2} \text{Log}_{\m{W}_l} (\m{X}) \cdot \right. \nonumber\\
&&  \left. \left\{ \begin{array}{lcc}
  \left(P(l|\m{X},y) - P(l|\m{X})\right) &\text{if} & c_l = y \\ 
  (- P(l|\m{X})  ) &\text{if} & c_l \neq y
\end{array}\right. \right]
\end{eqnarray}
where $0<\alpha<1$ is the learning rate. Apparently, the factor $P(l|\m{X},y) - P(l|\m{X})$ is always positive. 
In fact, $$ P(l|\m{X},y) - P(l|\m{X}) = \frac{P(l) e^{f(\m{X}, \m{W}_l)} \sum_{\{j:c_j \neq y\}}  P(j)e^{f(\m{X}, \m{W}_j) }}{ \left(\sum_{\{j:c_j = y\}}  P(j)e^{f(\m{X}, \m{W}_j) }\right) \left( \sum_{j=1}^M  P(j)e^{f(\m{X}, \m{W}_j) } \right) }$$ 
%
%The learning rule reflects the fact that prototypes with the same label as that of the data point are attracted to the data point, while prototypes with different labels from the data point are repelled.
%
Consequently, analogous to the LVQ in Euclidean spaces, wrong prototypes are pushed away from $\bm{X}$, while correct prototypes are dragged closer to $\m{X}$, to the extend to which prototype $\m{W}_l$ ``stand out" for $\m{X}$, among the correct prototypes.
Since we have a probabilistic formulation, all prototypes are moved, instead of the closest ones to X only, but to the extent they are ``relevant to" $\m{X}$.
%
%
%$$P_y(l|\m{X}) = \frac{e^{ -\delta^2(\m{X}, \m{W}_l)/2\sigma^2}}{\sum_{\{j:c_j = y\}}  e^{ \delta^2(\m{X}, \m{W}_j)/2\sigma^2 }}$$ 
%
%$$P(l|\m{X}) = \frac{e^{-\delta^2(\m{X}, \m{W}_l)/2\sigma^2}}{\sum_{j=1}^M  e^{- \delta^2(\m{X}, \m{W}_j)/2\sigma^2 }}$$

%  In this paper, we let the learning rate monotonically decreasing with time as in  \cite{Schneider2009,Fouad2012} :
%\begin{eqnarray}
%\alpha(t) &=& \frac{\alpha(0)}{1+\nu\cdot(t-1)}
%\end{eqnarray}
%where $\nu>0$ is a parameter determining the decay speed,  $t$ denotes the number of training epochs.
%
The proposed probabilistic learning Riemannian space quantization algorithm is then summarized by Algorithm \ref{al:PRLVQ}.

\begin{algorithm}[!h]
\caption{Probabilistic learning Riemannian space quantization algorithm (PLRSQ) }
\label{al:PRLVQ}
\begin{algorithmic}[1]
\REQUIRE {$m$ training examples $(\m{X}_1,y_1),...,(\m{X}_m,y_m)$, where $\m{X}_i \in \mathbb{S}^+(n)$ and $y_i \in \{1,...,C\}$, a positive scalar $\sigma^2$, a small learning rate $\alpha$ }
\ENSURE $M$ labeled prototypes $({\bf W}_1,c_1),...,{(\bf W}_M,c_M) $, where ${\bf W}_i \in \mathbb{S}^+(n))$ and $c_i \in \{1,...,C\}$
\STATE  Initialize $\m{W}_i$ by the Riemannian mean of examples labeled by $c_i$ plus small random perturbation. 
 %$\mathfrak{m}_0 = I$ 
\WHILE {a stopping criterion is not reached}
\STATE Randomly select a training example $(\m{X}_i,y_i)$
\pt{
\FOR{$l = 1,...,M$}
\STATE Compute  $$P(l|\m{X}_i,y_i) =  \frac{e^{ -\delta^2(\m{X}_i, \m{W}_l)/2\sigma^2}}{\sum_{\{j:c_j = y_i\}}  e^{ \delta^2(\m{X}_i, \m{W}_j)/2\sigma^2 }}$$ and $$P(l|\m{X}_i) = \frac{e^{-\delta^2(\m{X}_i, \m{W}_l)/2\sigma^2}}{\sum_{j=1}^M  e^{- \delta^2(\m{X}_i, \m{W}_j)/2\sigma^2 }}$$
\ENDFOR
\FOR{$l = 1,...,M$}
\IF {$c_l=y_i$}
\STATE $V_l = \frac{\alpha}{\sigma^2} \left(P(l|\m{X}_i,y_i) - P(l|\m{X}_i))\right) \text{Log}_{{\bf W}_l} ({\bf X}_i)$
\ELSE
\STATE  $V_l = - \frac{\alpha}{\sigma^2}   p(l|\m{X}_i) \text{Log}_{{\bf W}_l} ({\bf X}_i)$
\ENDIF
\STATE ${\bf W}_l \leftarrow \text{Exp}_{{\bf W}_l} (V_l)$
\ENDFOR}
\ENDWHILE
\end{algorithmic}
\end{algorithm}

\section{Experiments}
\label{sec:exp}
In this section, we verify our proposed probabilistic learning Riemannian space quantization on both synthetic and real world data sets. Our proposed probabilistic learning Riemannian Space quantization (PLRSVQ) with (PLRSVQ-AN) and without (PLRSVQ-Const) annealing in variance  were examined. For PLRSVQ-Const,  $\sigma^2 = \sigma^2_{opt} $ was used, while for PLRSVQ-AN, following \citep{Seo2003Soft}, an annealing schedule for the scale (temperature) parameter $\sigma^2$ was used:  $\sigma^2(t) = \sigma^2(t-1)\cdot \beta(t)$, $\beta(t) = 
(\beta(t-1))^{1.1}$, where $\beta(0) = 0.99$ and $\sigma^2(0) = \sigma^2_{opt} $ is a tunable hyper-parameter.
The annealing was terminated when  $\sigma^2(t) <  \sigma^2_{opt} - 0.4$ \citep{Seo2003Soft}. The learning rates are continuously reduced \citep{Schneider2009} as
$
\alpha(t) = \frac{n \xi }{100} \cdot 0.01^{t/T},
%\alpha_{max}\cdot\left(\alpha_{min}/\alpha_{max}\right)^{t/T},
$ where $n$ is the rank of the SPD matrices, $\xi$ represents the number of prototypes per class,  $T$ denotes the number of sweeps through the training data, and $t=1,...,T$.
Since the classification data sets used in this study are balanced,
we set the class priors to $P(l) = \frac{1}{M}, l=1,...,M$.
    
\subsection{Artificial Data sets}
We first generated synthetic data set to verify our proposed approach. The instances are generating in  polar coordinates according to the following equations: 
\begin{equation}
\label{eq:genrateP}
\m{X} = \sum_j^{n} \lambda_j \bm{u}_j \bm{u}_j^T
\end{equation} 
where $\lambda_j$ represents the  $j$-th eigenvalue of $\m{X}$ and $ \bm{u}_j$ is the corresponding eigenvectors. Here, we choose $n = 10$.

We designed four sets of eigenvalues. The first set of eigenvalues is from a linearly decreasing function:

\begin{eqnarray*}
\tilde{\eta}^1(j) = 13- j, \ \ j=1,...,n,  
\end{eqnarray*}

\noindent
The second set of eigenvalues follows an exponentially decreasing function:

\begin{eqnarray*}
\tilde{\eta}^2(j) = 1+100\exp( - 0.5 j), j=1,...,n,
\end{eqnarray*}

\noindent
The third set of eigenvalues is also follows from a linearly decreasing function but with different slope:

\begin{eqnarray*}
\tilde{\eta}^3(j) = 13- 0.5j, \ \ j=1,...,n,  
\end{eqnarray*}

\noindent
The fourth set of eigenvalues follows from a reciprocal function:

\begin{eqnarray*}
\tilde{\eta}^4(t) = \frac{1}{j}, \ \ j=1,...,n,  
\end{eqnarray*}

\noindent
For all four sets of eigenvalues, the mean of the eigenvalues is normalized to 1, i. e. 

$${\eta}^i (t)  = n\tilde{\eta}^i (t) / {\sum_{q=1}^n \tilde{\eta}^i (q)} $$
 
\noindent
where $i = 1,...,4$.

\begin{figure}[!h]
\centering
\includegraphics[scale=0.5]{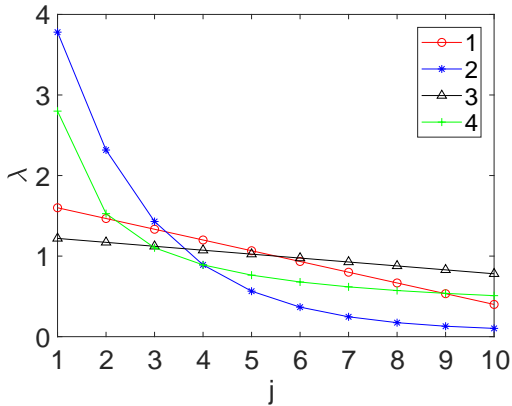}
\caption{Plot of the four sets of eigenvalues. 
%The values marked by red circle are eignvalues from the linearly decreasing function, while the values marked by blue * are the exponentially decreasing function.
}
\label{fig:lambdas}
\end{figure}

The four sets of eigenvalues are plotted in Fig.~\ref{fig:lambdas}. 

We designed two sets of eigenvectors (basis). To that end, we generated two $n \times n$ random real matrices (each element generated i.i.d. from $\mathcal{N}(0,1)$. Then, for each matrix, the Gram-Schmidt orthogonalization was used to obtain the orthogonal basis - the set of eigenvectors. We denote the two sets of eigenvectors as $\{\bm{v}^i_1,...,\bm{v}^i_n\}, i= 1,2$ respectively.

Two synthetic datasets were generated, named as SynI, SynII respectively. For SynI, the first two sets of eigenvalues and the two sets of eigenvectors are combined to produce four classes of instances. The first two classes share the first set of eigenvalues, but each with different sets of eigenvectors. The remaining two classes share the second set of eigenvalues, and also each with different sets of eigenvectors. When instances were generated,  random noise were injected in both its eigenvalues and eigenvectors. The eigenvalues $\lambda_j, j=1,...,n$ of the instance were created according to uniform distribution $U({\xi}^1(j)-\epsilon,{\xi}^1(j)+\epsilon )$ or $U({\eta}^2(j)-\epsilon,{\eta}^2(j)+\epsilon )$, depending on its class label. Here we chose $\epsilon = 0.1$. The eigenvectors ${\bm{u}}_j, j=1,...,n$ of the instance were the orthogonalized version of $\bm{v}^1_j + \bm{\epsilon}$ or $\bm{v}^2_j + \bm{\epsilon}$ through Gram-Schmidt orthogonalization depending on its label,  where $\bm{\epsilon}$ follows $ \mathcal{N}(0,{\nu}^2\m{I})$. Here we chose $\nu=0.3$.  Once, the eigenvalues and eigenvectors of the instance were obtained, the instance can be acquired by Eq.~\eqref{eq:genrateP}.

The SynII  is also of four classes and was generated using the four sets of eigenvalues and first set of eigenvectors. Instances of each class has its own set of eigenvalues but share the common eigenvectors. Each instance was created following the same procedure of SynI.

The synthetic datasets are summarized in Table \ref{tab:descri_syndata}.
For both the SynI and \pt{SynII}, a training set, a validation set, and a test set were generated independently. All three sets contained $250$ instances per class. The generating process of each dataset was repeated for 30 times, the following results are the average results over the 30 runs.  

\begin{table}[!h]
\centering
\caption{Descriptions of synthetic datasets. $l$ and $k$ denotes the number of sets of eigenvalues  and eigenvectors that are used to generate the data, respectively. $C$ denotes the number of classes, $n$ denotes the rank of the SPD matrices, $\#$Train represents the number of training instances,$\#$Validation denotes the number of validation instances, while $\#$Test is the number of test instances. }
\label{tab:descri_syndata}
\resizebox{0.45\textwidth}{!}{
\begin{tabular}{llllllll}\hline
Name & $l$ & $k$ & $C$ & $n$ & $\#$Train & $\#$Validation & $\#$Test \\\hline
SynI %\multirow{3}{*}{ SynI } 
&2&2&4& $10$ & $250*4$ & $250*4$ &  $250*4$ \\ 
SynII% \multirow{3}{*}{SynII} 
&4&1&4&  $10$ & $250*4$ & $250*4$&  $250*4$ \\  
           \hline 
\end{tabular}}
\end{table}

\subsubsection{Hyper-parameter Sensitivity Analysis}

\begin{figure}[h]
\centering
\subfloat[SynI dataset]{\includegraphics[width=0.35\textwidth]{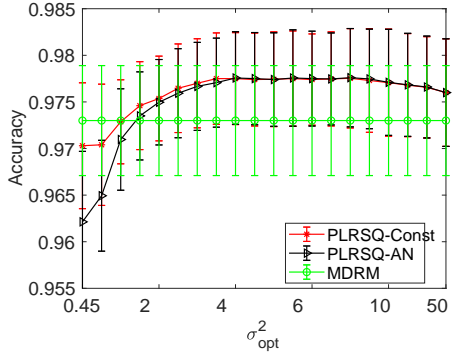}}
%\subfloat[Example of Changes of costs SynI]{\includegraphics[scale=0.3]{./fig/fixedsigmaCostSynI.eps}} \\

\subfloat[SynII dataset]{\includegraphics[width=0.35\textwidth]{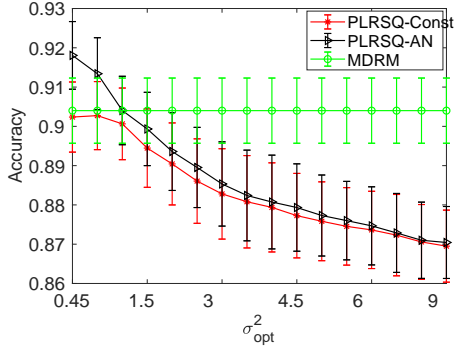}}
%\subfloat[SynII]{\includegraphics[scale=0.65]{./fig/annealingsigmaCostSynII.eps}}
\caption{ The averaged performance over 30 runs together with the standard derivation as a function of $\sigma^2_{opt}$ on the synthetic data set  }
\label{fig:Ressyn}
\end{figure}

To understand the stability of the PLRSQ method with respect to the hyperparameter  $\sigma^2_{opt}$, we show in Fig. \ref{fig:Ressyn}, the averaged classification rates with standard derivations across 30 runs on test sets, as a function of $\sigma^2_{opt}$. One prototype per class was used. The models were trained for 100 epochs. 
%
%We first explore the sensitivity of hyper-parameter $\sigma^2_{opt}$ in the proposed algorithm. We let $\sigma^2_{opt}$  $\in$ $\{0.45$, $0.5$,  $1$, $1.5$, $2$, $2.5$, $3$, $3.5$, $4$, $4.5$, $5$, $5.5$, $6$, $7$, $8$, $9$,  $10$, $20$, $30$, $50 \}$. The averaged performance over 30 runs together with the standard derivation on the test set as a function of $\sigma^2_{opt}$ is depicted in Fig. \ref{fig:Ressyn}, under the experimental setting of one prototype per class and 100 training epochs.  
%
From Fig. \ref{fig:Ressyn} we can see that the proposed PLRSQ method is sensitive to the parameter of $\sigma^2$. With well tuned $\sigma^2$, both PLRSQ-AN and PLRSQ-Const can outperform the MDRM algorithm, while PLRSQ-AN shows slightly better performance than PLRSQ-Const. This also suggests that the PLRSQ algorithm is different from MDRM even with one prototype per class. Instead of learning class centers like MDRM does, PLRSQ learns the decision boundary. 

% performance changed

\subsubsection{\pt{Investigation on the Impact of Learning Rate}}

\pt{As mentioned in the beginning of this section, we used the decaying learning rate $\alpha(t) = \frac{n \xi }{100} \cdot 0.01^{t/T}$. 
It is worth noting that 
the logarithm and exponential maps are only accurate in a sufficiently small neighborhood on the manifold.
To see the impact of the initial learning rate on our algorithm, we increased and decreased $\alpha(0)$ to $\frac{n \xi }{50}$ and 
$\frac{n \xi }{200}$. The results\footnote{We thank one of the anonymous reviewers for the suggestion.} are given in Table~\ref{tab:difflearnrates}. 
To remove the impact of $\sigma^2$, we set $\sigma^2 = 1.5$ (this value yielded acceptable performance for both synthetic data sets). Again, one prototype per class was used. The model was trained for 100 epochs. Averaged performance over 30 trials together with standard derivation is reported. Table~\ref{tab:difflearnrates} reveals that these changes in the learning rate did not lead to any dramatic difference.

\begin{table}
\centering
\caption{Averaged accuracy over 30 trials together with standard derivations under different learning rates }
\label{tab:difflearnrates}
\begin{tabular}{llll} \hline
Learning rate   & $n \xi \setminus 50$   & $n \xi \setminus 100$  & $n \xi \setminus 200$\\\hline
SynI & 97.50 $\pm$ 0.46 & 97.43 $\pm$ 0.43 & 97.41 $\pm$ 0.47  \\
SynII & 89.78 $\pm$ 0.86 & 89.49 $\pm$ 0.92 & 88.78 $\pm$ 1.00
\\ \hline
\end{tabular}
\end{table}

As we decreased the learning rate, the performance on the SynII data set slightly degenerated. This could be because we kept training the model for 100 epochs though we used a smaller learning rate. We then increased the number of training epochs to 150 for learning rate  $\alpha(t) = \frac{n \xi }{200} \cdot 0.01^{t/T}$, the performance of our method on the SynII data set increased to an average accuracy of 89.23 ($\pm$ 0.98).  Our method will eventually converge as long as the learning rate (in annealing schedule) is small. However, if we used a smaller learning rate, we need to train the model longer to obtain the same performance. Thus, a balance between the choice of the learning rate and the training time need to be considered. Our future work will consider self-adapted learning rate. 
}

\subsubsection{\pt{Investigation on Convergence}}

\begin{figure}[!h]
\centering
\subfloat[Cost on SynI]{\includegraphics[width=0.25\textwidth]{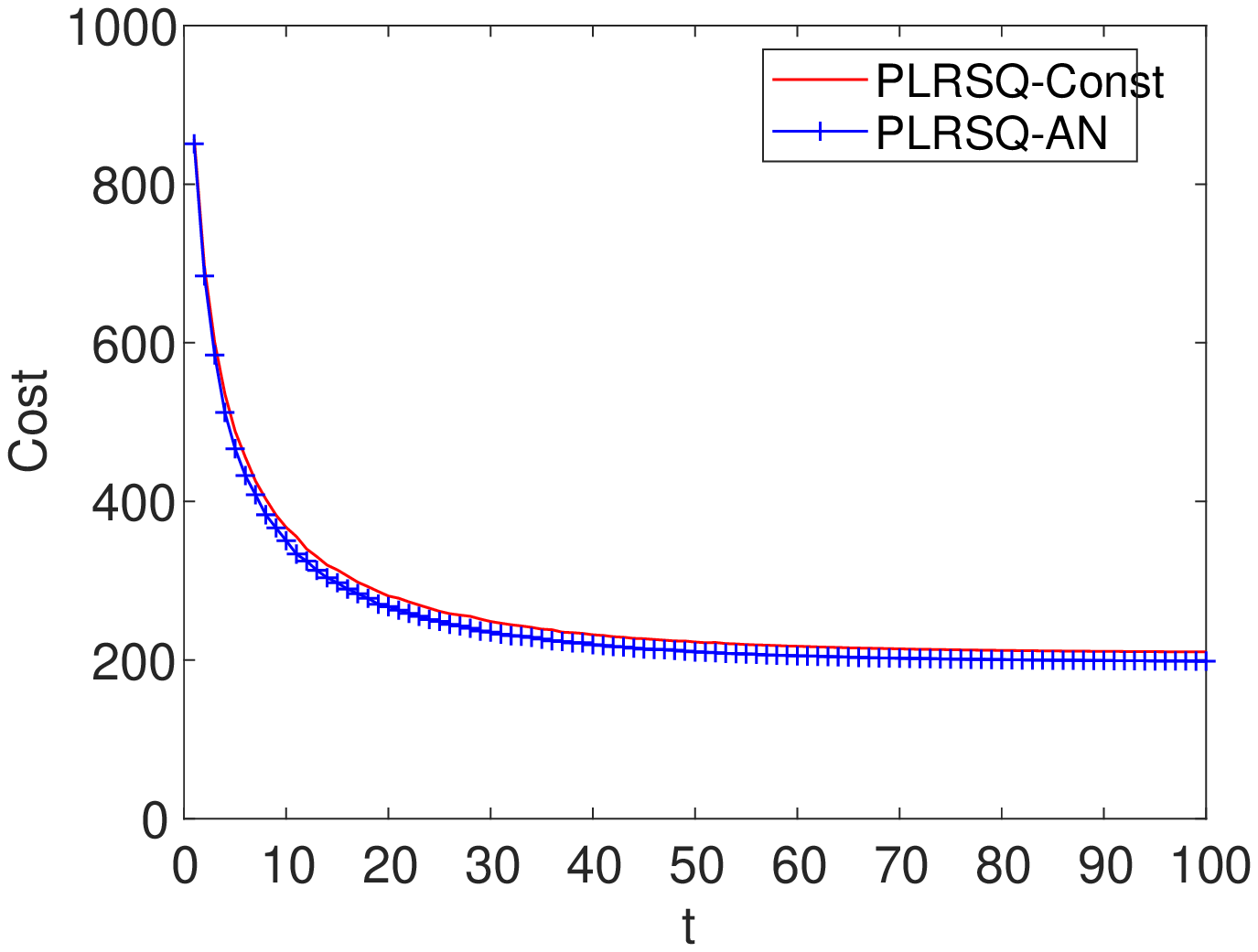}}
\subfloat[Cost on SynII]{\includegraphics[width=0.25\textwidth]{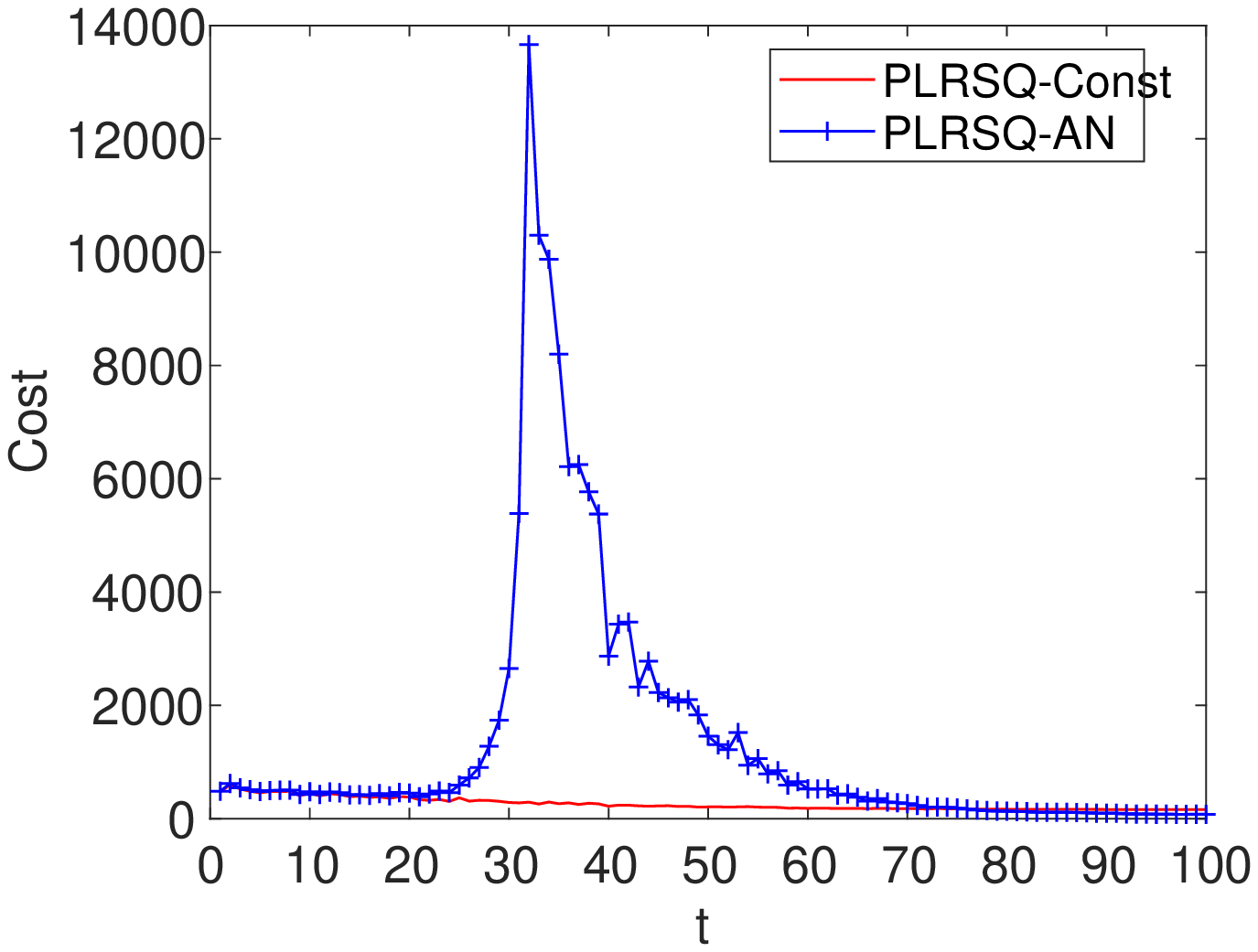}}\\
\subfloat[Training error on SynI]{\includegraphics[width=0.25\textwidth]{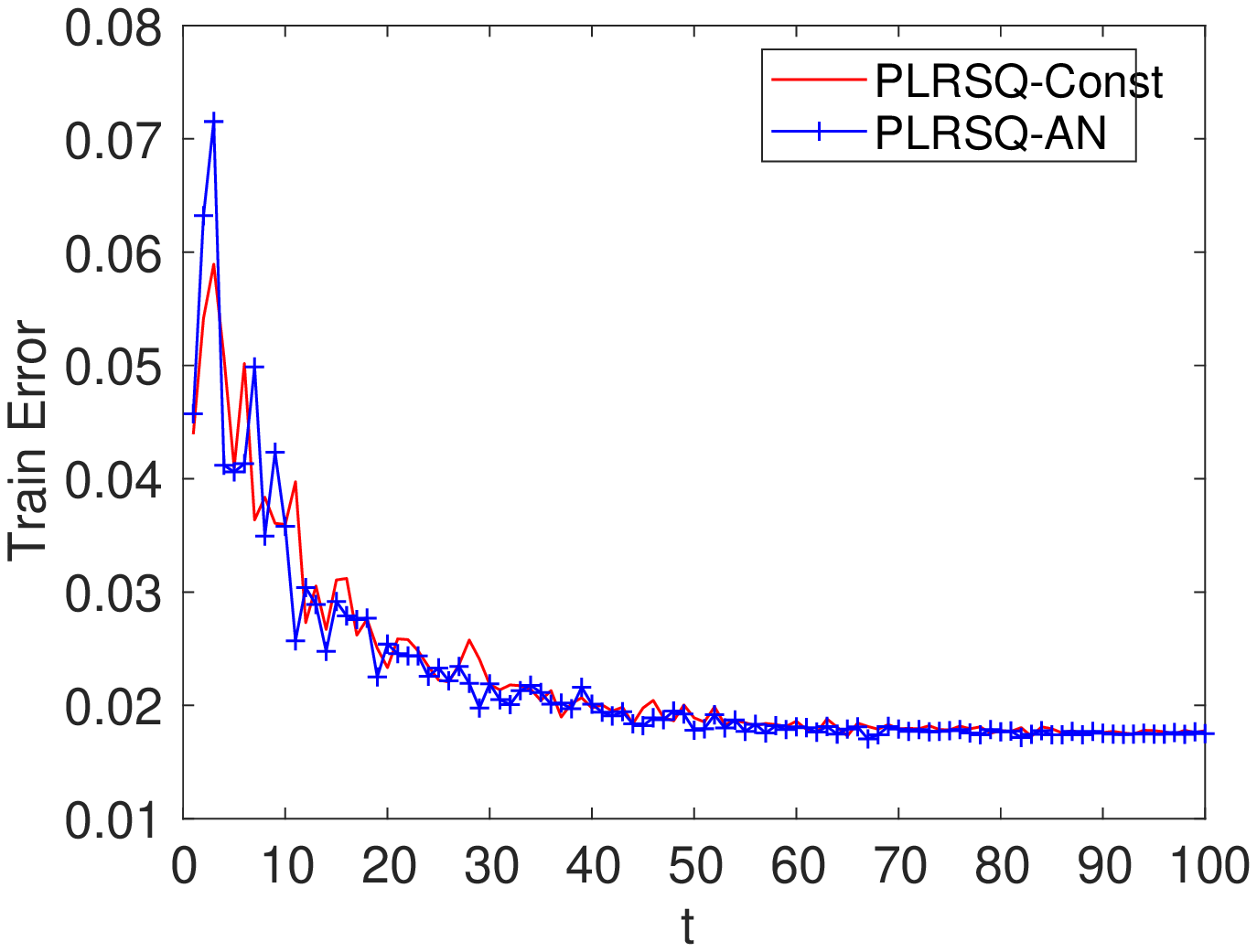}}
\subfloat[Training error on SynII]{\includegraphics[width=0.25\textwidth]{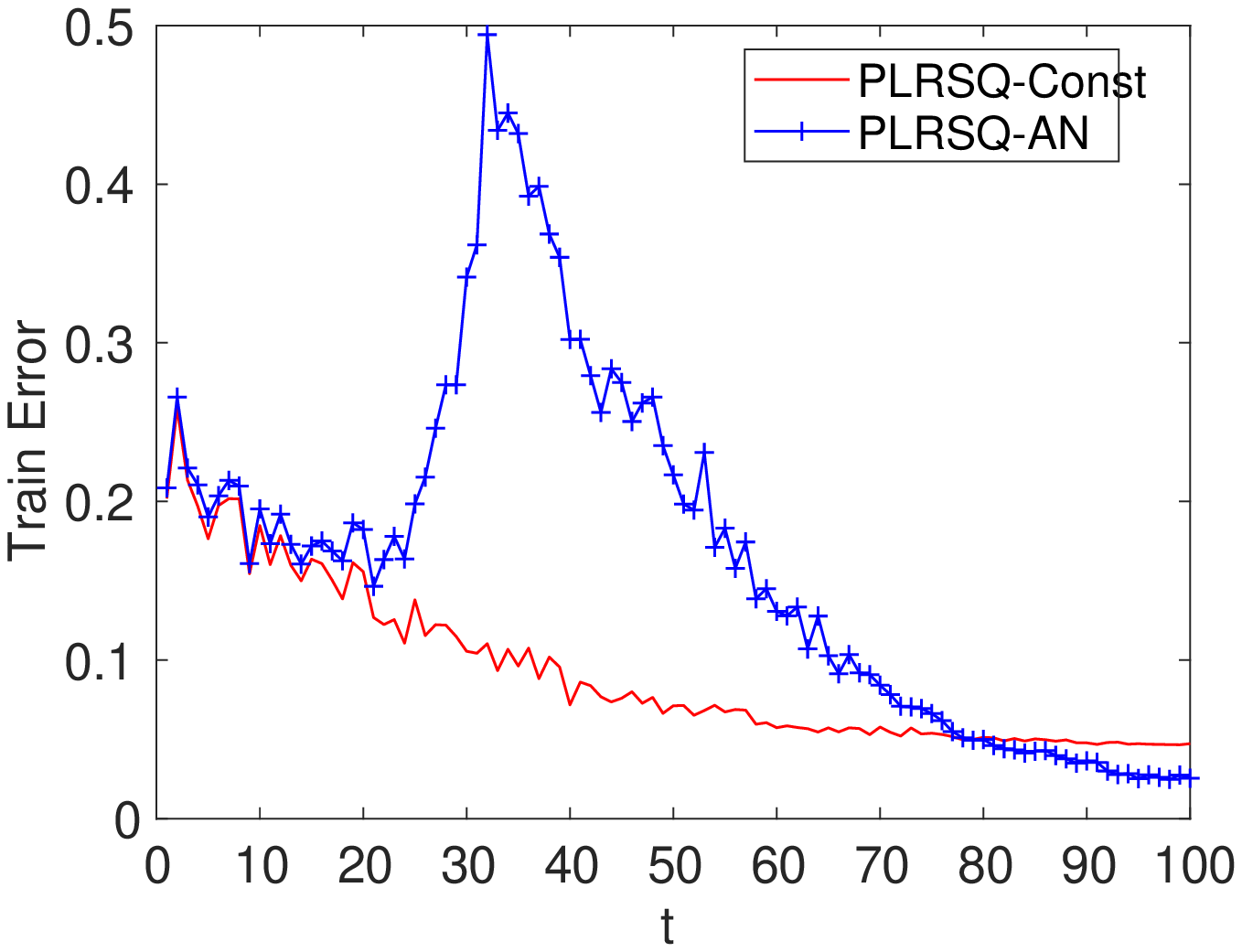}}\\
\subfloat[Test error on SynI]{\includegraphics[width=0.25\textwidth]{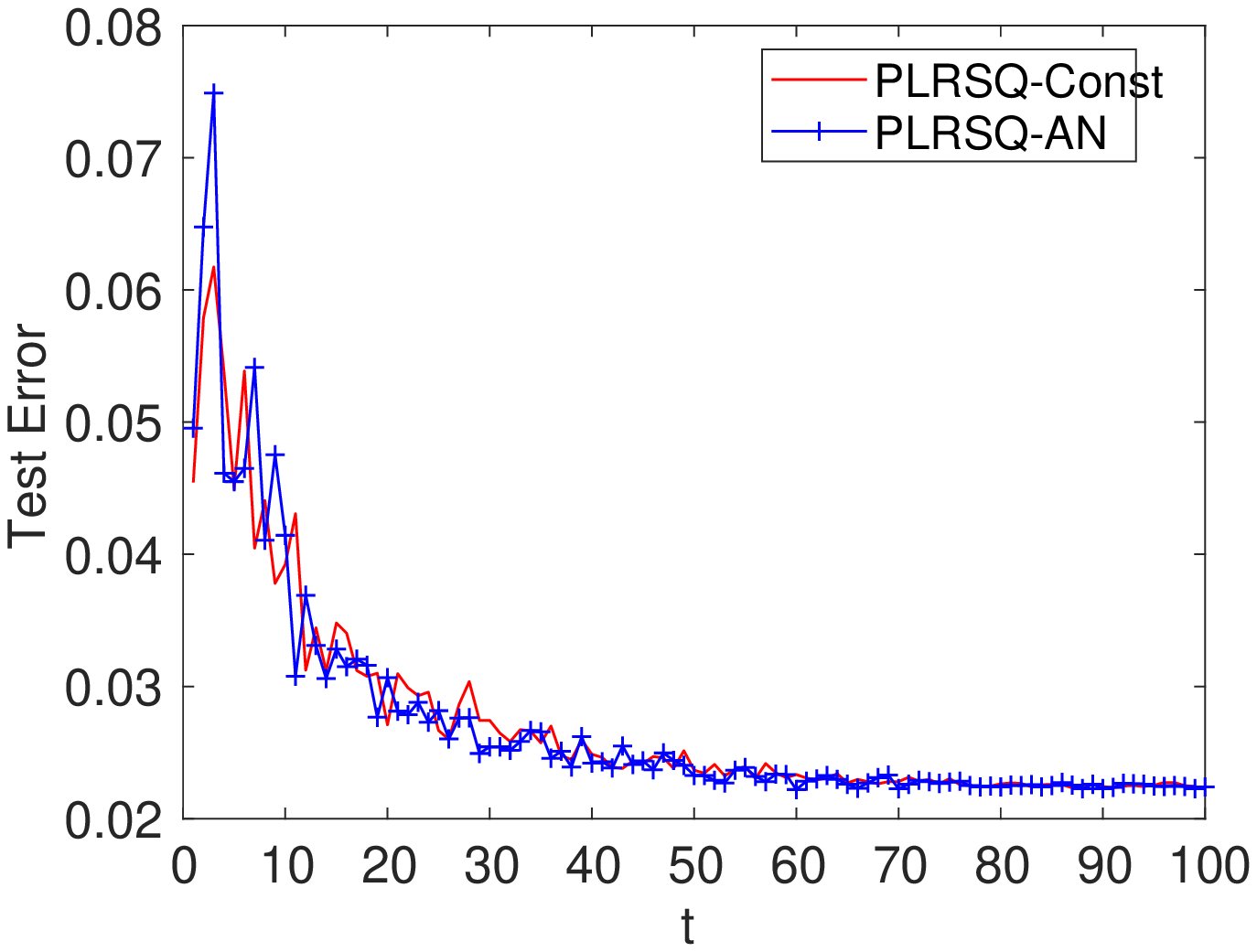}}
\subfloat[Test error on SynII]{\includegraphics[width=0.25\textwidth]{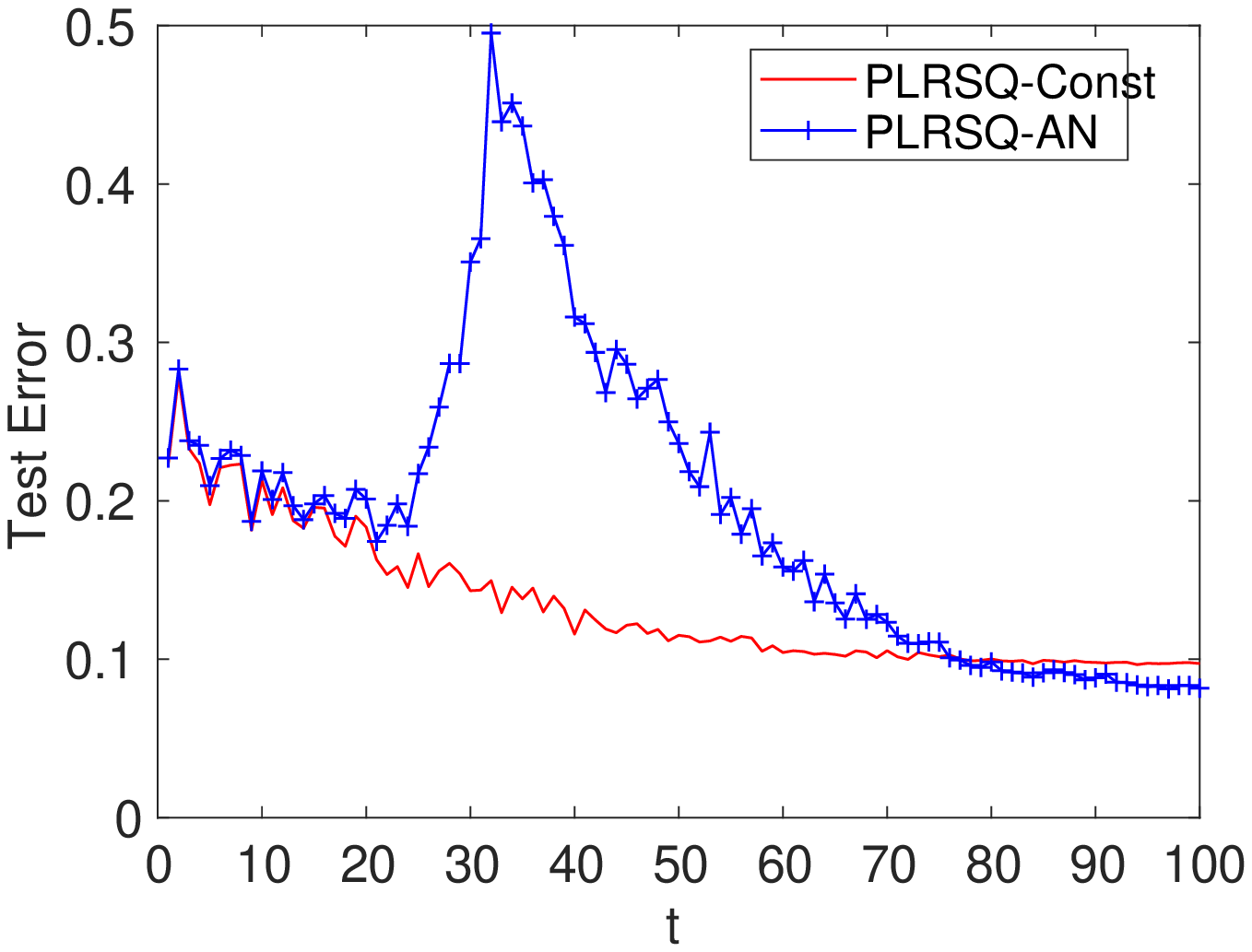}}
\caption{ Performance change during training course on data sets SynI and SynII .  }
\label{fig:ConvSynI}
\end{figure}

To see the convergence property of the PLRSQ method, we show in Fig. \ref{fig:ConvSynI}, the  evolution of cost function, training error, and test error during the training course of the models. The models were trained for 100 epochs. We selected $\sigma^2_{opt} = 8$ for both PLRSQ-Const and PLRSQ-AN on dataset SynI,  while $\sigma^2_{opt} = 0.5$ for PLRSQ-Const and $\sigma^2_{opt} = 0.45$ for PLRSQ-AN on dataset SynII, as models with these values produced best classification rates on test sets.  

% performance changed

As given by Fig. \ref{fig:ConvSynI} (a), (c) and (e), for data set SynI, the PLRSQ-Const and PLRSQ-AN show similar behavior during the training course.  The costs of both PLRSQ-Const and PLRSQ-AN decrease monotonically. The PLRSQ-AN converges to slightly smaller value compared to PLRSQ-Const.  The training error of both PLRSQ-Const and PLRSQ-AN slightly fluctuate and eventually converge to a similar stable value. The same trend is found for test error. However, for data set SynII, the behavior of PLRSQ-Const and PLRSQ-AN is completely different. The PLRSQ-Const shows nice monotonic convergence learning curve, but the PLRSQ-AN gives non-monotonic learning curve. However, the PLRSQ-AN converge to smaller values compared to PLRSQ-Const for cost, training error, and test error on data set SynII. Overall, the PLRSQ-AN algorithm shows better performance compared to PLRSQ-Const, but may deliver non-monotonic learning curve on some data set. 

\pt{The non-monotonic learning curve appears to be caused by the heuristic annealing schedule of $\sigma^2$. The time evolution of $\sigma^2$ during training is shown in Figure \ref{fig:ConvSynII-sigma}. It initially changes very fast. When using a slower decay schedule by fixing $\beta$ to 0.99 (see Figure~\ref{fig:ConvSynII-sigma-slower}), we can obtain a naturally converging learning curve (Figures \ref{fig:ConvSynII-cost-slower} and \ref{fig:ConvSynII-testError-slower}). This is in agreement with the findings in \cite{Schneider2010_hpl} that heuristic annealing schedule of $\sigma^2$ may lead to non-monotonic learning curves in RSLVQ. In our future work, we will consider a systematic treatment of $\sigma^2$, instead of the current heuristic schedule. 
\begin{figure}
\centering
\subfloat[Changes of $\sigma^2$]{\includegraphics[width=0.25\textwidth]{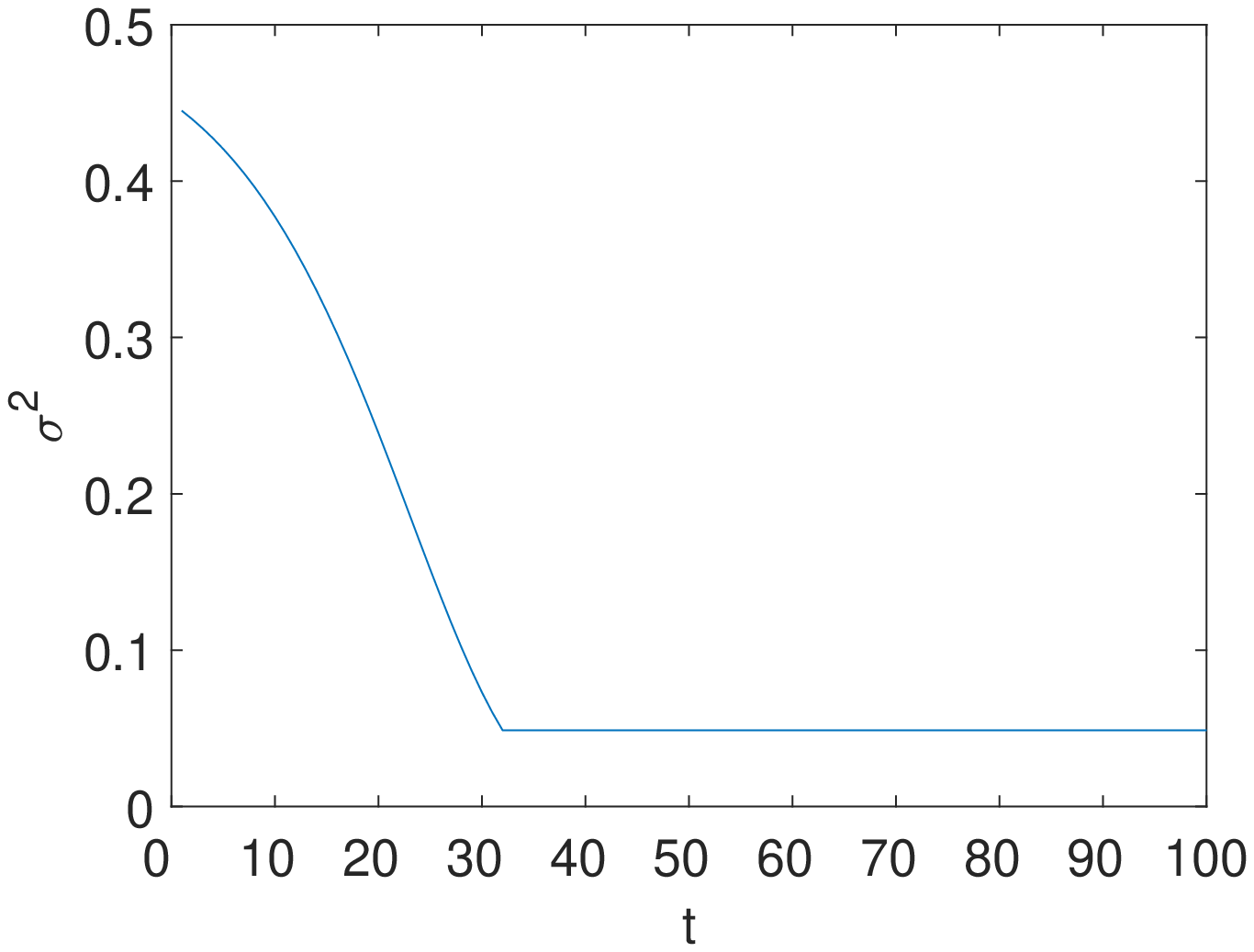} \label{fig:ConvSynII-sigma}}
\subfloat[Slower changes of $\sigma^2$]{\includegraphics[width=0.25\textwidth]{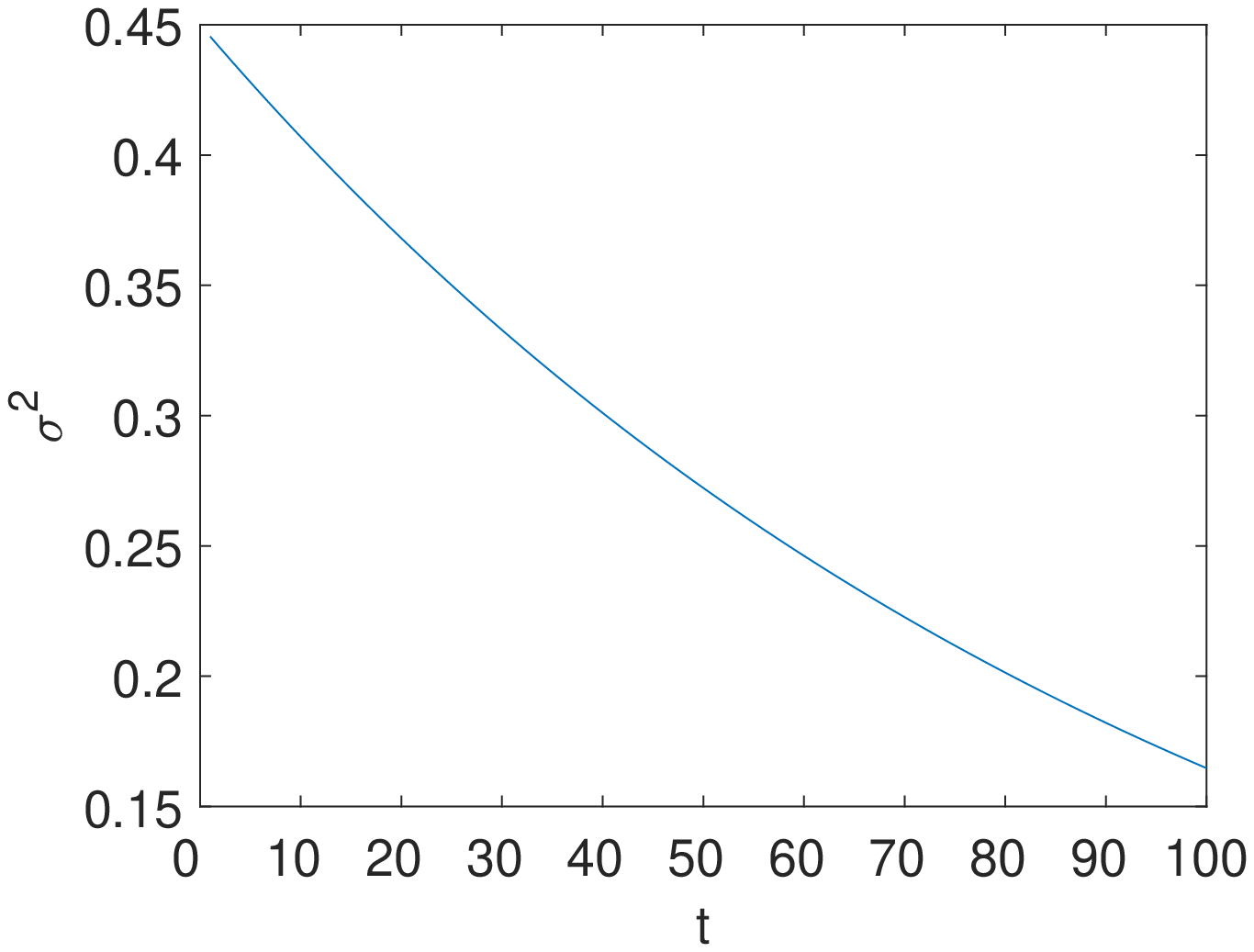}\label{fig:ConvSynII-sigma-slower}}\\
\subfloat[Costs with slower changes of $\sigma^2$]{\includegraphics[width=0.25\textwidth]{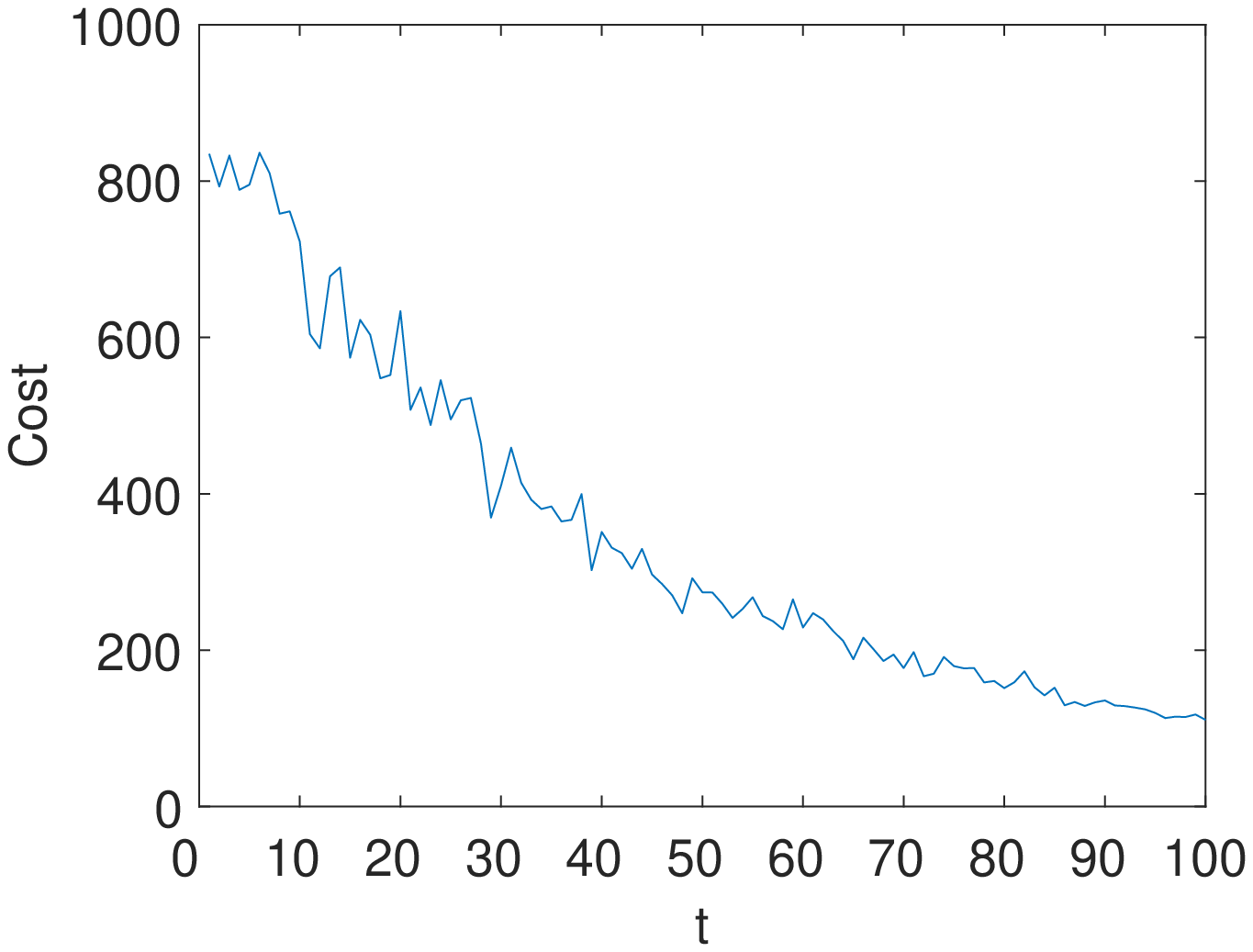}\label{fig:ConvSynII-cost-slower}}
\subfloat[Test error with slower changes of $\sigma^2$]{\includegraphics[width=0.25\textwidth]{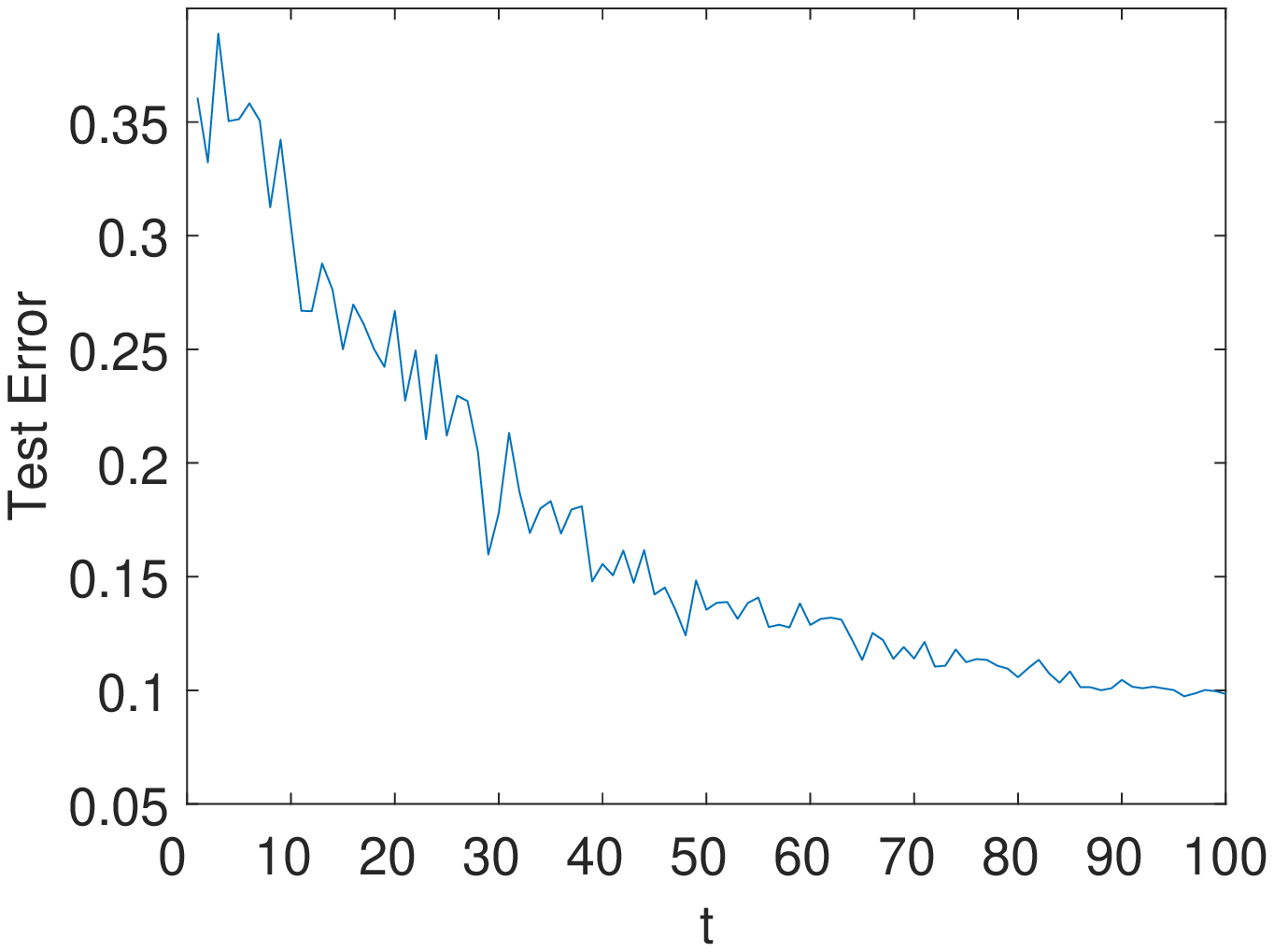}\label{fig:ConvSynII-testError-slower}}\\
\caption{Performance change during training course on SynII with slower changes of $\sigma^2$ }
\label{fig:ConvSynII-slower}
\end{figure}

}

% As given by Fig. \ref{fig:ConvSynI}, for data set SynI, the costs of both PLRSQ-Const and PLRSQ-AN decrease monotonically and the PLRSQ-AN converges to slightly smaller value compared to PLRSQ-Const; the training error and test error of both PLRSQ-Const and PLRSQ-AN eventually converge to similar stable values.  

% % performance changed
%\begin{figure*}[!h]
%\centering
%\subfloat[Change of cost]{\includegraphics[scale=0.35]{./fig/synIICost.eps}}\ 
%\subfloat[Change of training error]{\includegraphics[scale=0.35]{./fig/synIItrainError.eps}}\
%\subfloat[Change of test error]{\includegraphics[scale=0.35]{./fig/synIItestError.eps}}\
%\caption{ Performance change during training course on data set SynII.  }
%\label{fig:ConvSynII}
%\end{figure*}

\subsubsection{Performance Comparison}

We compared the final performance of our proposed method with the MDRM method. We trained the PLRSQ algorithms on the training set, and test the algorithms using validation set and test set, respectively. The hyperparameter $\sigma^2_{opt}$ were selected from 0.45 to 50 based on the validation set performance. The number of prototypes per class and training epochs were selected from $\{1,2,3\}$, and $\{20,50,100\}$, respectively, also based on the validation set performance. The test performance with optimal hyper-parameters was reported. As the MDRM method consists no tuning parameters, to provide a fair comparison, we trained MDRM with both training and validation sets and tested it using test set.

%\begin{table}[!h]
%\caption{Comparison of accuracy among different LVQ approaches on synthetic data sets. Averaged accuracy over 30 runs along with standard deviation is given.}\label{tab:rSyn}
%\begin{tabular*}{\tblwidth}{@{} LLL@{} }
%\toprule
%Method & SynI & SynII \\
%\midrule
%MDRM & 0.9737 $\pm$ 0.0053 & 0.9074 $\pm$ 0.0089\\ 
%PLRSQ-Const& 0.9765 $\pm$ 0.0052 & 0.9051 $\pm$ 0.0107\\ 
%PLRSQ-AN &  \bf{0.9770 $\pm$ 0.0055}  & \bf{0.9211 $\pm$ 0.0073}\\ \hline
%\bottomrule
%\end{tabular*}
%\end{table}

\begin{table}[!h]
\centering
\caption{Comparison of accuracy between our method and MDRM on synthetic data sets. Averaged accuracy over 30 runs along with standard deviation is given. }
\label{tab:rSyn}
 \resizebox{0.45\textwidth}{!}{
\begin{tabular}{lcc} \hline
%\toprule
Method & SynI & SynII \\ \hline
%\midrule 
%PLVQ-Const & &\\
%PLVQ-AN &0.9407 $\pm$ 0.0083& \\
MDRM & 0.9737 $\pm$ 0.0053 & 0.9074 $\pm$ 0.0089\\
%GMLVQ & 0.9361 $\pm$ 0.0078 & 0.8191 $\pm$ 0.0155 \\
PLRSQ-Const& 0.9765 $\pm$ 0.0052 & 0.9051 $\pm$ 0.0107\\ 
PLRSQ-AN &  \bf{0.9770 $\pm$ 0.0055}  & \bf{0.9211 $\pm$ 0.0073}\\ 
 \hline
%\bottomrule
\end{tabular}}
\end{table}

The test set classification performance (averaged accuracy over 30 runs with standard deviation) on synthetic datasets SynI and \pt{SynII} is presented in Table \ref{tab:rSyn}. {The optimal parameters (i.e. $\sigma_{opt}$, number of prototypes per class, and training epochs) are listed in the Table \ref{tab:paraSyn} in \ref{app:OptPar}.} For data set SynI, PLRSQ with (PLRSQ-An) and without (PLRSQ-Const) annealing in variance preforms significantly better than the MDRM method ($p=2.35 \times 10^{-6}$ and $p=4.21 \times 10^{-5}$ via non-parametric Wilcoxon signed-rank test \citep{Wilcoxon1945}, respectively). The performance of PLRSQ with annealing in variance on data set SynI shows slightly better performance than that without annealing in variance ($p = 0.24$). For data set SynII, PLRSQ with annealing in variance preforms significantly better than the MDRM method ( $p=1.71 \times 10^{-6}$) and the PLRSQ without annealing in variance ( $p = 3.48 \times 10^{-6}$), while there is no significant difference between the MDRM method and PLRSQ without annealing in variance ($p = 0.48$).

Since PLRSQ with annealing in variance shows better generalization performance, we used this version of PLRSQ in the following experiments.

\subsubsection{Nonlinear Riemannian Structure Verification}
 
To see the impact of the nonlinear Riemannian structure, we compared the proposed PLRSQ method with the baseline \pt{robust soft} learning vector quantization (\pt{RSLVQ}) \citep{Seo2003Soft} method that utilizes Euclidean distance to measure the distance between the manifold-valued instance and the corresponding prototype, totally ignoring the nonlinear structure. 
\pt{The updated prototypes were restricted to be positive definite as follows: If, following the eigen-decomposition of $\m{W}$, negative eigenvalues were set to 0, we would obtain the projection (in the $L_2$ sense) of $\m{W}$ onto the space of symmetric positive semi-definite matrices. We impose a small threshold $\tau>0$ and replace all negative eigenvalues of $\m{W}$, as well as the eigenvalues smaller than $\tau$, by $\tau$. In this way the prototypes are projected onto the space of SPD matrices. Hence, the method can be viewed as a kind of projected gradient descent. In the experiments we used $\tau=10^{-4}$.
} 

Following PLRSQ, the annealing in variance was used in RSLVQ and the corresponding hyper-parameters were tuned using validation data sets. Selected parameters were given in Table \ref{tab:paraSyn}.

\pt{
It may be also interesting to compare our method with generalized matrix learning vector quantization (GMLVQ) \citep{hammer2002} or generalized tangent learning vector quantization\footnote{We are thankful to one of the anonymous reviewers for this suggestion.} \citep{Saralajew2016}.  Such methods automatically learn adaptive Mahalobis distances between data points and class prototypes, providing optimal class discrimination. Learning of the metric is purely driven by the classification task. The learnt metric usually points to a low-dimensional discriminatory subspace. The data manifold itself is accounted for only implicitly and exclusively through the data sample. In cases of high-dimensional data, the manifold structure cannot be accounted for easily when the samples are relatively small. In our case, we know the Riemannian manifold structure of SPD matrices and take the full use of it when formulating PLRSQ. Of course, one can argue that from the performance perspective, a good out-of-sample performance may be satisfactory, even though the learnt metric is driven by considerations other than accounting for the data manifold accurately. We thus provide a performance comparison between our method and GMLVQ/RSLVQ to assess the usefulness of explicitly taking the Riemannian manifold structure of SPD matrices into an account (see Table~\ref{tab:comREsyn}). The number of prototypes per class and training epochs in GMLVQ were also selected from $\{1,2,3\}$ and $\{20,30,100\}$, respectively, based on the validation set performance.  Selected parameters were given in Table \ref{tab:paraSyn}.
GMLVQ cannot be directly used for data points of SPD matrices, as it can not preserve the positive definiteness of the considered SPD matrices. We therefore again apply the projection procedure described above for RSLVQ to the updated prototypes. The results are given in Table~\ref{tab:comREsyn}. 

%Our method is different from generalized matrix learning vector quntization (GMLVQ) \citep{hammer2002} or generalized tangent learning vector quantization \citep{Saralajew2016}.  These method automatically learn adaptive Mahalobis or tangent distance between data points and class prototypes, providing the best class discrimination. They assumed that the inherent data manifold dimensionality is much lower than that of the emebdding data space. The data manifold is considered implicitly and exclusively based on the data sample. In contrast, we know the Riemannian manifold structure of SPD matrices and take the full use of it when formulating PLRSQ. Nevertheless, we provided performance comparison between our method and GMLVQ to see the usefulness of explicitly taking the Riemannian manifold structure of SPD matrices into consideration (see Table~\ref{tab:comREsyn}). GMLVQ can not be directly used for data points of SPD matrices, as it can not preserve the positive definiteness of the considered SPD matrices. We therefore apply the projection procedure of section to the updated prototypes. The results are given in Table~\ref{tab:comREsyn}.
}

%Table.~\ref{tab:comREsyn} compares the performance of the probabilistic learning vector quntization using Riemannian geodesic distance (PLRSQ) with that of using Euclidean distance (RSLVQ)  in the space of SPD matrices on synthetic data sets.

%\begin{figure}[!h]
%\centering
%\includegraphics[width=0.3\textwidth]{./fig/comREsyn.eps} 
%\caption{Performance comparison between probabilistic learning vector quntization using Riemannian geodesic distance (denoted by PLRSQ in the figure) and Euclidean distance (denoted by PLVQ) in the space of SPD matrices on synthetic data sets.  
%\label{fig:comREsyn}}
%\end{figure}

\begin{table}[!h]
\centering
\caption{Performance comparison between probabilistic learning vector quantization using Riemannian geodesic distance (PLRSQ), Euclidean distance (RSLVQ), and Mahalobis distance (GMLVQ) in the space of SPD matrices on synthetic data sets. }
\label{tab:comREsyn}
 \resizebox{0.45\textwidth}{!}{
\begin{tabular}{lcc} \hline
%\toprule
Method & SynI & SynII \\ \hline
%\midrule 
%PLVQ-Const & &\\
%PLVQ-AN &0.9407 $\pm$ 0.0083& \\
%MDRM & 0.9737 $\pm$ 0.0053 & 0.9074 $\pm$ 0.0089\\
RSLVQ& 0.9407 $\pm$ 0.0083 & 0.7752 $\pm$ 0.0125\\ 
GMLVQ & 0.9361 $\pm$ 0.0078 & 0.8191 $\pm$ 0.0155 \\
PLRSQ &  \bf{0.9770 $\pm$ 0.0055}  & \bf{0.9211 $\pm$ 0.0073}\\ 
 \hline
%\bottomrule
\end{tabular}}
\end{table}

\pt{Table~\ref{tab:comREsyn} suggests that the method using Riemannian geodesic distance (PLRSQ) can significantly outperform RSLVQ and GMLVQ  on both synthetic data sets in terms of classification accuracy. 
Exploiting the nonlinear Riemannian structure of the data manifold through Riemannian geodesic distance can indeed lead to improved classification performance. }

%From Table~\ref{tab:comREsyn}, we can see that the method using Riemannian geodesic distance significantly outperformed that using flat Euclidean distance on both synthetic data sets in terms of classification accuracy. The results indicates that exploiting the nonlinear Riemannian structure of the data points through Riemannian geodesic distance leads to improved classification performance.   

\subsection{Image Classification}
The ETH-80 dataset contains 8 categories with 10 objects each and 41 images per object.  Following \citep{Jayasumana2015}, we used 21 randomly chosen images from each object to train the classifier and the rest to evaluate the out-of-sample classification accuracy. For each image, we used a single $5 \times 5$ covariance descriptor calculated from the features $[x, y, I, |I_x|, |I_y|]$, where $x$, $y$ are pixel locations and $I$, $I_x$, and $I_y$ are corresponding intensity and derivatives \citep{Jayasumana2015}. The split of training and test set was randomly and independently repeated  20 times. 
%We report the mean test set classification rates along the std dev across 20 splits.
The models were trained for 100 epochs. 

% \begin{figure}[!h]
%\centering
%\includegraphics[width=0.45\textwidth]{./fig/eth80dataPlot.eps}
%\caption{Sample images of ETH-80 dataset.}
%\label{fig:ResETH80data}
%\end{figure}

\begin{figure}[!h]
\centering
\subfloat[Classification Performance of the proposed method for different $\sigma^2_{opt}$ with 2 prototype per class. Different marked curves correspond to tasks of identifying different number of categories, e. g. the curve $3C$ denotes as that of identifying 3 categories of the images.]{\includegraphics[width=0.35\textwidth]{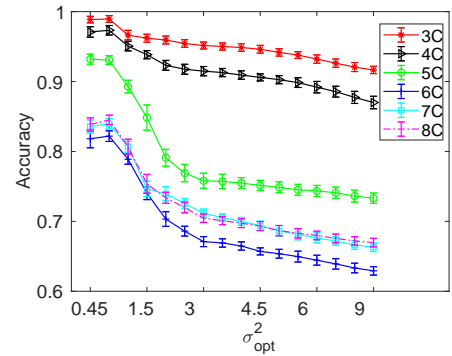} \label{fig:ResETH80Par}} \\
\subfloat[Classification accuracy as a function of the number of prototypes per class for identifying different number of categories in the ETH80 dataset.]{\includegraphics[width=0.35\textwidth]{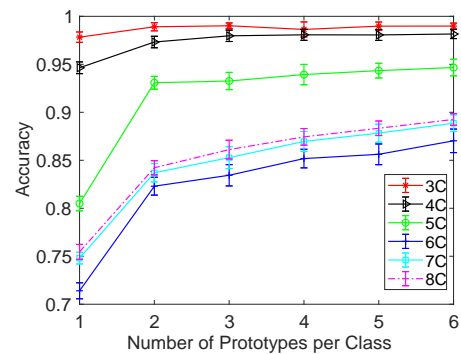} \label{fig:ResETH80Nopro}}
\caption{Classification performance on  ETH80 dataset under different settings. %a) Performance of the proposed method for different $\sigma^2_{opt}$ with 1 prototype per class. $3C$ denotes as identifying 3 categories of the images. b) Classification accuracy as a function of the number of prototypes per class for identifying different number of categories in the ETH80 dataset.
}
\end{figure}

To understand the stability the PLRSQ method with respect to the hyperparameter $\sigma^2_{opt}$ in classification tasks with increasing number of classes, we show in Fig. \ref{fig:ResETH80Par} the mean classification rates with standard deviations across 20 splits, as a function $\sigma^2_{opt}$.
Two prototypes per class was used. The six curves correspond to performances on tasks with 3, 4, ...,8 classes. The classes were ordered as follows: {\it apple, car, cow, cup, dog, horse, pear} and {\it   tomato}.
Fig. \ref{fig:ResETH80Nopro} presents classification rates of PLRSQ as a function of the number of prototypes per class. The value of $\sigma^2_{opt}$ was set through 5-fold cross-validation using  values ranging from 0.45 to 50.
Both figures indicate a decreasing performance trend as more categories are involved in the classification task. Interestingly, identifying 6 categories (denoted as 6C in the Fig. \ref{fig:ResETH80Par} and Fig. \ref{fig:ResETH80Nopro}) yielded worse performance than identifying 7 or 8 categories. One possible reason is that {\it dog}, {\it horse}, and {\it cow} are rather difficult to distinguish. 
However,the last two classes {\it pear} and {\it tomato} are more easily distinguishable from the other classes. This is collaborated by the performance drop of MDRM on 6 categories (see  Table \ref{tab:eth80resultsw}).

%% performance changed
% \begin{figure}[!h]
%\centering
%\includegraphics[width=0.35\textwidth]{./fig/eth80Par.eps}
%\caption{Performance of the proposed method for different $\sigma^2_{opt}$ with 1 prototype per class. $3C$ denotes as identifying 3 categories of the images.}
%\label{fig:ResETH80Par}
%\end{figure}
%
%
%
%% performance changed
% \begin{figure}[!h]
%\centering
%\includegraphics[width=0.35\textwidth]{./fig/eth80Re6pros.eps}%{./fig/eth80NoProt.eps}
%\caption{Classification accuracy as a function of the number of prototypes per class for identifying different number of categories in the ETH80 dataset.  }
%\label{fig:ResETH80Nopro}
%\end{figure}

%\pt{ Fig.~\ref{fig:comRE} demonstrates the performance of PLRSQ compared to that of PLVQ on the ETH-80 data set. From Fig.~\ref{fig:comRE}, we can see that the method using the Riemannian geodesic distance in the space of SPD matrices significantly outperformed that using the Euclidean distance.  
%
%\begin{figure}[!h]
%\centering
%\includegraphics[width=0.4\textwidth]{./fig/comREeth80.eps} 
%\caption{Performance comparison between probabilistic learning vector quntization using Riemannian geodesic distance (denoted by R in the figure) and Euclidean distance (denoted by E) in the space of SPD matrices on the ETH-80 data set.  
%\label{fig:comRE}}
%\end{figure}
%
%}  

%The number of prototypes per class, the training epochs, and $\sigma^2_{opt}$ were selected from $\{1,2,3,4\}$, $\{50,100,200\}$ and 
Table \ref{tab:eth80resultsw} compares the performance of PLRSQ and MDRM with Affine-Invariant metric with that of 
k-means(KM) and kernel k-means (KKM) on Riemannian manifolds with Log-Euclidean metric
on the ETH80 data set (results taken from \cite{Jayasumana2015}, only mean classification rates were reported). For 
PLRSQ and MDRM we provide the mean performances and standard deviations across the 20 data splits.
Compared with all three methods,
PLRSQ achieved a superior performance.

\begin{table}[!h]
\centering
\caption{Comparison of our proposed method with k-means(KM) and kernel k-means (KKM) on Riemannian manifolds with Log-Euclidean metric.}
\label{tab:eth80resultsw}
\begin{tabular}{llllll}\hline
$C$ & KM  & KKM & MDRM  & PLRSQ  \\ \hline
3  & 75.00 & 94.83 & 88.76 $\pm$ 0.75& \bf{98.95} $\pm$ 0.55\\  
4  & 73.00 & 87.50 & 83.61 $\pm$ 0.60& \bf{98.10} $\pm$ 0.53\\ 
5  & 74.60 & 85.90 & 68.85 $\pm$ 0.77& \bf{94.27} $\pm$ 0.74\\ 
6  & 66.50 & 74.50 & 60.03 $\pm$ 0.65& \bf{86.59} $\pm$ 1.41\\  
7  & 59.64 & 73.14 & 63.46 $\pm$ 0.49& \bf{88.56} $\pm$ 1.04\\ 
8  & 58.31 & 71.44 & 63.16 $\pm$ 0.60& \bf{89.09} $\pm$ 0.95\\ \hline
\end{tabular}
\end{table}

{Admittedly, on image classification tasks, deep convolutional neural networks (CNN) and related approaches can implicitly learn to handle the underlying structure of the input space and no doubt can achieve performances that are difficult to beat. However, they need large samples for robust learning of the multitude of free parameters. Instead, our proposed PLRSQ method can learn on comparatively small samples, since the model capacity can be low and is easily controlled by the number of prototypes used for each class. To illustrate the robustness of our method in the case of reduced sample sizes,  {a large image data set CIFAR-10 \citep{Krizhevsky2009} containing 50000 training and 10000 test images organized in 10 classes was downsampled in a controlled manner. \pt{In particular, we randomly drew 200, 500, 1000, 5000, 10000, and 20000 images from the training set (in a stratified manner) to train our method, as well as the VGG net \citep{Simonyan2015}, a representative deep CNN method.} The methods were then verified on the hold-out set of 10000 test images. The random downsampling of training images was repeated 5 times. We report the average performance together with standard derivation across the 5 runs}.

\pt{As for the VGG network, the 11 layer structure was used (detailed configuration is given by A in Table 1 in \citep{Krizhevsky2009}). Each network was trained for 100 epochs. Each VGG network was trained for 100 epochs. Random initialization, as well as using VGG pretrained on the ImageNet dataset were considered. 
Note that the pretraining gives VGG an advantage since pre-training involved presentation of many other training instances and thus useful hints of prior knowledge that PLRSQ cannot have. Nevertheless, we included the pretrained VGG to judge to what extent can pretraining on the ImageNet data help the more complex VGG model cope with limited sample sizes.
For random initialization, the learning rate was annealed according to the exponential decay schedule, i.e. $\alpha(t) = 0.01 * 0.99^t$. For pre-trained VGG, a smaller learning rate $\alpha(t) = 0.001 * 0.99^t$ was found to be preferable since the network needed to perform fine tuning to the new data set only.  As for our method, the number of prototypes per class and the hyperparameter $\sigma_{opt}^2$ were chosen from 1 to 5, and 0.45 to 5 , respectively, via 5-fold cross validation on training set.%
%
%Both random initialization and smart initialization by model pretrained on the ImageNet dataset were considered. For random initialization,  the learning rate was annealed according to the exponential decay schedule, i.e. $\alpha(t) = 0.01 * 0.99^t$. For pre-trained VGG, a smaller learning rate $\alpha(t) = 0.001 * 0.99^t$ was used. } As for our method, the number of prototypes per class and the hyperparameter $\sigma_{opt}^2$ were chosen from 1 to 5, and 0.45 to 5 , respectively, via 5-fold cross validation on training set.
} 
Following \citep{Vemulapalli2015Riemannian}, each image is represented by  $9 \times 9$ covariance descriptor calculated from the features $[x, y, R, G, B, |I_x|, |I_y|, |I_{xx}|, |I_{yy}|]$, where $x$, $y$ are pixel locations and $I$, $I_x$, and $I_y$ are corresponding intensity and derivatives, $I_{xx}$ and $I_{yy}$ are corresponding second order partial derivative \footnote{We have tried using $5 \times 5$ covariance descriptor. The performance of our method gives better performance when using $9 \times 9$ covariance descriptor.}.

\pt{The mean test accuracy curves (together with $\pm$ standard deviation bars) are shown as functions of the number of training examples in Fig. \ref{fig:comRDL}. When the training sample is small, our method performed  better than the randomly initialized VGG net, but, as expected, our method consistently performed worse than the pre-trained VGG net.} 

%The mean test accuracy as a function of the number of training examples are demonstrated in Fig. \ref{fig:comRDL}. \pt{As we can see in Fig. \ref{fig:comRDL}, during the situations when the size of training samples is small, our method performed  better than the VGG net, but our method consistently performed worse than the pre-trained VGG net. Our method only used the current training instances. However, the pre-trained VGG net has used prior knowledges existing in the very large ImageNet dataset.
 
\pt{Moreover, compared with the VGG net, our method is much more computationally efficient. 
%The model size of VGG is much larger than our method. 
The VGG net considered in this paper has 133 million free parameters, while our method operates with at most 4051 parameters (max 5 prototypes per class). Training times per epoch of the VGG net and our model (in the most complex case of 5 prototypes per class) are presented in Figure.~\ref{fig:comRDL_time}. The VGG net was implemented in Python, our method was coded in Matlab. The two methods were run using the same computer without GPU. The test time of the VGG net on the entire set of 10000 instances was roughly 26.3 seconds, while the test time of our method (5 prototypes per class) was less than half that (approximately 10.5 seconds).
}

%Nevertheless, our method is much computational efficient than the VGG net. The model size of VGG is much larger than our method. The VGG net involved in this paper contains 133 million parameters, while our method contains 4051 parameters at most (5 prototypes per class at most). The training time per epoch of the VGG net and our method at worse case (5 prototypes per class) are given by Figure.~\ref{fig:comRDL_time}. The VGG net was implemented using Python, our method was realized using Matlab. The two method were run using the same computer without GPU. The test time of the VGG net on the entire 10000 instances is roughly 26.3 seconds, while the test time of our method (5 prototypes per class) on the entire 10000 instances is only around 10.5 seconds.

\pt{The training time of our method depends linearly on the number of training instances. For each training instance, the method needs to compute its Riemannian distance to all prototypes. 
% Thus, the computational time of our algorithm  also depends linearly on the number of prototypes. 
The calculation of the Riemannian distance involves eigenvalue decomposition. The time complexity of eigenvalue decomposition for an $n \times n$ matrix is (at most) $O(n^3)$.  Thus, the time complexity of our algorithm is $O(Mmn^3)$, where $M$ represents the number of prototypes, $m$ denotes the number of training instances, and $n$ is the rank of training instances (SPD matrices).}

%The training time of our method depends linearly on the number of training instances. For each training instance, the method needs to compute its Riemannian distance to all prototypes. Thus, the computational time of our algorithm  also depends linearly on the number of prototypes. The calculation of the Riemannian distance involves . The time complexity of eigenvalue decomposition for an $n \times n$ matrix is $O(n^3)$ .  Thus, the time complexity of our algorithm is $O(Mmn^3)$, where $M$ represents the number of prototypes, $m$ denotes the number of training instances, and $n$ is the rank of training instances (SPD matrices). 

\begin{figure}[!h]
\centering
\includegraphics[width=0.4\textwidth]{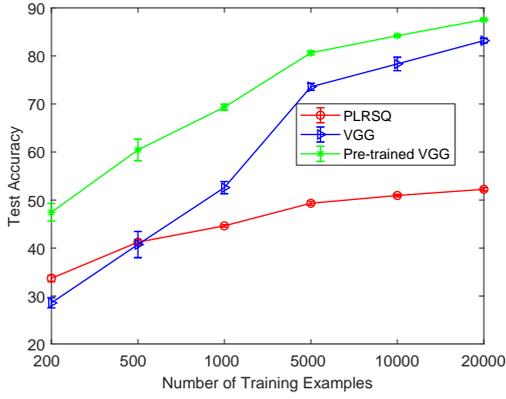} 
\caption{Out-of-sample accuracy as a function of the number of training examples for the PLRSQ method and both randomly initialized and pretrained VGG.
\label{fig:comRDL}}
\end{figure}

\begin{figure}[!h]
\centering
\includegraphics[width=0.4\textwidth]{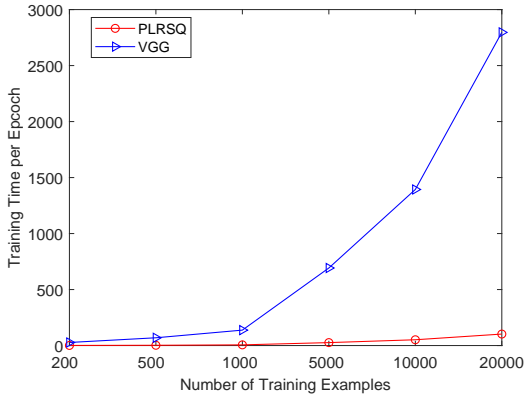} 
\caption{Training time per epoch changes as a function of the number of training examples.
\label{fig:comRDL_time}}
\end{figure}

%\begin{table}
%\centering
%\caption{Comparison between our method and VGG  in terms of selected model size and computational time the on training set of size 200}
%\label{tab:modelcomp}
%\begin{tabular}{lll} \hline
%Method & VGG & PLRSQ \\ \hline
%No. of parameters & 133 M & 811\\
%Training time/epoch (GPU)& 14.5 s & -\\
%Training time/epoch (CPU)& & 15.9 s\\
%Test time (GPU)& 2.3 s & -\\ 
%Test time (CPU)& & 4.4s\\ 
%\hline
%\end{tabular}
%\end{table}

% VGG-nets use a simple yet effective strategy of constructing very deep networks stacking building blocks of the same shape

\subsection{Motor Imagery Classification}

We also tested PLRSQ on the task of motor imagery classification
used in the BCI competition IV \citep{Tangermann2012Review}. In particular, we used data set 2a since this was the only dataset with non-binary motor imagery tasks.
We represented the electroencephalogram (EEG) recording during each motor imagery task through the  sample covariance matrix, summarizing spatial information in the signal with temporal content integrated out
\cite{Congedo2017}.

The data set 2a in the BCI competition IV  consists of EEG signals from 9 healthy subjects who were performing four different motor imagery tasks, i.e. imagination of the movement of the left hand, right hand, both feet and tongue. The signals were recorded by placing 22 electrodes distributed over sensorimotor area of the subject at a sampling rate of 250 Hz. For each subject, two sessions were recorded on two different days, each containing 288 trials with 72 trials per class (one day constituted the training data, the other out-of-sample test set). At each trial, a cue was given in the form of an arrow pointing either to the left, right, down or up, corresponding to one of the four classes, to prompt the subject to perform the corresponding motor imagery task. The motor imagination lasted 4 seconds from the presence of cue till the end of motor imagery task. The time interval of the processed data was restricted to the time segment comprised between 0.5 and 2.5 s starting from the cue instructing the user to perform the mental task. EEG signals from each trial were bandpass filtered by a 5-th order Butterworth filter in the 10-30 Hz frequency band. The filtered signal were zero mean. Then pre-processed EEG signals were transformed into spatial covariance matrices (the hypothesis is that for the given tasks, the spatial covariance matrix provides sufficient discriminative information about the brain states). Suppose the $i$-th trial of  pre-processed EEG signal is given as follows:
\begin{equation}
\bm{E}_i = [ \bm{e}(t_i),...,\bm{e}(t_i+l-1)] \in \mathbb{R}^{n \times l}
\end{equation}
where $n$ and $l$ denote the number of channels and sampled points, respectively. %For motor imagery EEG data, the spatial covariance matrix provides discriminative information for brain states. Thus in this paper, 
\noindent
Each trial of EEG signal is represented by the sample covariance matrix computed as follows:
\begin{equation}
\label{eq:EEG2P}
\m{X}_i  = \frac{1}{l-1} \m{E}_i\m{E}^T_i
\end{equation} 
\noindent
Thus, elements of the IV 2a data set live in $\mathbb{S}^{+}(22)$.

We compared our method with the-state-of-the-art methods in the literature for motor imagery EEG decoding as listed in Table \ref{tab:rBCIIV2atrte}. The methods labeled by \textbf{1$^{st}$}, \textbf{2$^{nd}$}, and \textbf{3$^{rd}$} are the first three winning methods in BCI competition IV  on the data set 2a \citep{Tangermann2012Review}. These approaches are based on the common spatial patterns (CSP) method \citep{Moritz2008Multiclass}, or its variant--the filter bank common spatial pattern \citep{Ang2008filter}.
{Common spatial patterns is a widely used feature extraction method in  motor imagery based EEG classification. It is based on the assumption that  neural activation are spatially distributed in cortex areas and changes in the variance of EEG data in specific frequency bands indicate the intention of the user. Given EEG data of two different classes, the CSP algorithm computes the spatial filters that maximize the variance ratio of the data conditioned on two classes. 
\textbf{Wavelet-spatial Convolutional Network (WaSF ConvNet)} \cite{Zhao2019} is a deep learning approach that learns joint space-time-frequency features through using Morlet wavelet-like kernels and spatial kernels. Since deep networks contain many free parameters, the data was enlarged by cropped training. Furthermore, subject-to-subject weight transfer was used.
\textbf{Tangent Space Linear Discriminate Analysis (TSLDA)} \citep{Barachant2012Multiclass}, as mentioned in the introduction, map SPD matrices onto the tangent space at their Riemannian geometric mean and then apply standard linear discriminate analysis in the tangent space. \textbf{Tangent Space of Sub-manifold with Linear Discriminate Analysis (TSSM+LDA)}\citep{Xie2017Motor} learns an optimal map from the original Riemannian space of SPD matrices to the Riemannian sub-manifold through jointly diagonalizing the Riemannian means of the two class data. The optimal map transforms the original SPD matrices into lower dimensional SPD matrices where tangent space linear discriminate analysis is then applied.        

\begin{table*}
\centering
\caption{Performance comparison between our method and the state-of-the-art methods in terms of kappa value on BCI competition IV dataset 2a. }
\label{tab:rBCIIV2atrte}
 \resizebox{0.9\textwidth}{!}{
\begin{tabular}{lllllllllll} \hline
Method & mean kappa & S1 & S2 & S3 & S4 & S5 & S6 & S7 & S8 & S9 \\ \hline
TSSM+LDA & \bf{0.59} & {\bf 0.77}& 0.33 & {0.77} & \emph{0.51} & 0.35 & 0.36 & \emph{0.71} & 0.72 & \bf{0.83}  \\
PLRSQ & \bf{0.59} & 0.75 & 0.34 & \bf {0.80} & \bf{0.58} & 0.38 & 0.37 & 0.70 & 0.64 & 0.75  \\
%\bf{RGLVQ} & \emph{0.586} &\bf{0.79}&	0.32 & \emph{0.76} & \bf{0.55} &0.34 &\emph{0.36} & 0.66 &0.70 &\emph{0.79}  \\\hline
WaSF ConvNet  & 0.58 & 0.63 & 0.32 & 0.75 & 0.44 & \bf{0.60} & \bf{0.38} & 0.69 & 0.71 & 0.73 \\
1$^{st}$ & {0.57} & 0.68 & \bf{0.42} & 0.75 & 0.48 & \emph{0.40}& 0.27 & \bf{0.77} & \bf{0.75}& 0.61 \\
TSLDA & 0.57 & 0.74& 0.38 & 0.72 & 0.50 & 0.26 & 0.34 & 0.69 & 0.71 & 0.76 \\ 
MDRM  & 0.52 & {0.75} & 0.37 & 0.66 & {0.53} & 0.29 & 0.27& 0.56 & 0.58 & 0.68 \\ 
2$^{nd}$ &0.52 & 0.69 & 0.34 & 0.71 & 0.44 & 0.16 & 0.21 & 0.66 & 0.73 & 0.69 \\
%GLVQ& 0.35 &0.50 &0.16& 0.50& 0.31 & 0.13 & 0.25 &0.33 &0.51&	0.44 \\\hline
3$^{rd}$ &0.31 & 0.38 & 0.18 & 0.48 & 0.33 & 0.07 & 0.14 & 0. 29 & 0. 49 & 0.44\\ \hline
%GMLVQ &0.31& 0.40&0.15 &0.36 &0.29&0.06&0.20&0.33&0.50&0.55 \\\hline
%GRLVQ & 0.24 & 0.40 &0.08 &0.28 &0.16 &0.06 &0.15&0.31&	0.42&0.34  \\\hline
\end{tabular}}
\end{table*}

Table~\ref{tab:rBCIIV2atrte} presents performances of the competing approaches on 9 subjects (S1--S9) in terms of kappa values - a widely used performance metric in motor imagery classification tasks \citep{Zhao2019}.  For four balanced classes, $\kappa = (P_a - 1/4)/(1-1/4)$, where $P_a$ is the classification accuracy.
For our method, the heperparameters $\sigma^2_{opt}$ (ranged from 0.45 to 50), number of prototypes per class (1--4),  number of training epochs (20, 50, 100) were selected using 5-fold cross validation on the training folds. 

The PLRSQ method is among the three methods that can beat the other methods on 2 out of 9 subjects.  It outperforms the MDRM and TSLDA methods on 7 and 6 subjects, respectively.
PLRSQ also shows superior performance to the deep WaSF ConvNet method on 6 subjects.  One possible reason may be the relatively small size of the subject-specific training sample.
Our method obtained comparable performance to the TSSM+LDA method, which takes advantage of sub-manifold learning. This suggests that dimension reduction of the covariance matrices may lead to improved classification performance. Our future work will incorporate 
Riemannian manifold dimension reduction in our  probabilistic learning Riemannian space quantization framework.  However, we note that the TSSM+LDA method was specifically designed for 
motor imagery based EEG classification tasks, whereas our PLRSQ method is general for tasks involving SPD matrices.

\section{Conclusion}
\label{sec:con}
In many real machine learning applications, the input features are represented by symmetric positive-definite (SPD) matrices living on curved Riemannian manifolds, rather than vectors living in flat Euclidean spaces. In such cases, traditional learning algorithms may fail to account for the natural structure of the data space and consequently yield inferior performance.
In this paper, we have proposed a novel approach for the classification of SPD matrices. It is a generalization of probabilistic learning quantization to the Riemannian manifold of SPD matrices equipped with affine-invariant metric. By defining the conditional probability of assigning a label to a given SPD matrix using the notion of Riemannian geodesic distance, we obtained prototype update rules via minimizing the negative log likelihood using stochastic Riemannian gradient descent algorithm.  

Our proposed  probabilistic learning Riemannian space quantization (PLRSQ) has 
inherited the intuitive and simple nature of learning vector quantization methods. Furthermore, it can naturally deal with multi-class classification of SPD matrices. More importantly, the approach is an online learning algorithm, which is able to perform life-long learning and the test is very fast, as it only needs to compare the test instance to several prototypes. Empirical experiments, conducted on two real world data sets, suggest a promising potential of our method. 

The proposed PLRSQ method is derived on the Riemannian manifold equipped with affine-invariant metric, empirically showing better performance than the K-means and kernel K-means methods derived on Riemannian manifold with Log-Euclidean metric. An interesting research question is whether the probabilistic learning vector quantization framework or the affine-invariant metric can outperform  the alternative Riemannian K-means or kernel K-means methods. Our future work will derive probabilistic learning vector quantization on the Riemannian manifold equipped with Log-Euclidean metric, providing a direct comparison to Riemannian K-means or kernel K-means methods  with Log-Euclidean metric. 

Our proposed PLRSQ is sensitive to the variance parameter $\sigma^2$. The $\sigma^2$ parameter was set manually in this paper. One of our future work will be devoted to automatically adapt the variance during the training course.

\begin{appendix}

\section{Calculation of Riemannian gradient of squared Riemannian distance}
\label{app:gradientcompute}
The Riemannian gradient of the squared Riemannian distance $\delta^2(\m{X}_i, \m{W}_{l})$ defined by Eq.\eqref{eq:Riemaniandistance}  can be computed through:
\begin{eqnarray}
\langle \m{V}_l, \nabla_{\m{W}_l} \delta^2_l \rangle_{\m{W}_l} = \frac{\text{d}}{\text{d}t} \delta^2(\m{X}_i, \gamma_l (t)) \bigg|_{t=0} \label{eq:atimeDe}
\end{eqnarray}
with the Riemannian inner product $\langle \cdot, \cdot \rangle_{\m{W}_l}$ defined by Eq.\eqref{eq:RieNatnmetric}, and geodesic curve $\gamma_l (t)$
defined by Eq.~\eqref{eq:geoCurve}.

The time derivative of the squared Riemannian distance along the geodesic curve is given as follows \citep{Moakher2008A,Thomas2004}:
\[
%\begin{eqnarray}
\frac{\text{d}}{\text{d}t} \delta^2(\m{X}, \gamma_l (t)) \bigg|_{t=0} 
= 2\text{Tr} \left[\m{V}_l  \log (\m{X}^{-1} \m{W}_l) \m{W}_l^{-1}\right].
\label{eq:tderiRieWJ}
%\end{eqnarray}
\]
Since $\log (\m{X}^{-1}) = - \log (\m{X})$,
we have
\[
\frac{\text{d}}{\text{d}t} \delta^2(\m{X}, \gamma_l (t)) \bigg|_{t=0} 
= - 2\text{Tr} \left[\m{V}_l \log ( \m{W}_l^{-1}\m{X}) \m{W}_l^{-1}\right]. 
\]
\noindent
Since $\m{W}_l^{-1} \m{W}_l = I$, where $I$ represents the identity matrix, we can then rewrite the above result as the inner product:

\begin{eqnarray*}
&& \frac{\text{d}}{\text{d}t} \delta^2(\m{X}, \gamma_{l} (t)) \bigg|_{t=0} \\
&=& -2 \text{Tr} (\m{V}_l \m{W}_l^{-1} \m{W}_l \log ( \m{W}_l^{-1}\m{X})\m{W}_l^{-1} )\\
&=&\langle\m{V}_l, -2\m{W}_l \log ( \m{W}_l^{-1}\m{X})  \rangle_{\m{W}_l}
\end{eqnarray*}
and from \eqref{eq:atimeDe} we deduce
\noindent
 $$\nabla_{\m{W}_l} \delta_l^2 = -2\m{W}_l \log ( \m{W}_l^{-1}\m{X}) .$$

\noindent
For any $A \in Gl(n)$, it holds $\log (A^{-1}BA) = A^{-1}(\log B) A$ \citep{Curtis1979}. Furthermore, since $\m{W}_l \in \mathbb{S}^+(n)$, $  \m{W}_l^{-1} =  \m{W}_l^{-1/2} \m{W}_l^{-1/2}$, the above equation can be rewritten as follows:
\begin{eqnarray}
\nabla_{\m{W}_l} \delta_l^2  &=& -2\m{W}_l \log ( \m{W}_l^{-1/2} \m{W}_l^{-1/2}\m{X}_i\m{W}_l^{-1/2} \m{W}_l^{1/2}) \nonumber\\
&=&  -  2 \m{W}_l^{1/2} \log (  \m{W}_l^{-1/2}\m{X}\m{W}_l^{-1/2} )\m{W}_l^{1/2} \nonumber\\ 
&=& - 2 \text{Log}_{\m{W}_l} (\m{X}) 
\end{eqnarray}

\section{Optimal Parameters}
\label{app:OptPar}

The optimal parameters used in the synthetic experiments are listed in Table \ref{tab:paraSyn}, the optimal parameters used in ETH80 are given by Table \ref{tab:paraETH80}, while those used in the motor imagery classification are given in Table \ref{tab:paramotordata}.  

\begin{table}[h!]
\caption{Selected number of prototypes per class,  training epochs, and $\sigma^2_{opt}$ on synthetic data sets}
\label{tab:paraSyn}
\centering
\begin{tabular}{lllllllll}\hline
    % & \multicolumn{2}{c|}{GLVQ} & \multicolumn{2}{c|}{GRLVQ} &\multicolumn{2}{c|}{GMLVQ} &\multicolumn{2}{c|}{RGLVQ} \\\hline
    %& \multicolumn{3}{c|}{PLRSQ-Const} & \multicolumn{3}{c|}{PLRSQ-AN} \\\hline
Method & Parameters & Syn I & SynII\\\hline
%\multirow{3}*{PLRSQ-Const} &  $\#$ prototype\\ 
%						   & $\#$epochs&  \\
%                           & $\sigma^2_{opt}$ & \\\hline 
%\multirow{3}*{PLRSQ-AN}    &  $\#$ prototype\\ 
%						   & $\#$epochs&  \\
%                           & $\sigma^2_{opt}$ & \\\hline 
\multirow{3}*{PLRSQ-Const}& $\#$ prototype & 2.57 $\pm$ 0.57  & 2.57 $\pm$ 0.62 \\
                              & $\#$epochs&  88.33 $\pm$ 21.51  & 80.67$\pm$ 25.38\\
                              & $\sigma^2_{opt}$ & 8.00 $\pm$ 7.24  &0.53 $\pm$ 0.16\\\hline 
\multirow{3}*{PLRSQ-AN}&
							   $\#$ prototype  & 2.43 $\pm$ 0.63  & 1.57 $\pm$ 0.73\\
                              & $\#$epochs & 95.00 $\pm$ 15.26 &75.00 $\pm$ 15.43\\
                              & $\sigma^2_{opt}$  & 8.82 $\pm$ 8.15  & 0.45 $\pm$ 0.01\\\hline  
\multirow{3}*{RSLVQ}    &  $\#$ prototype & 2.87 $\pm$ 0.35 & 1.87 $\pm$ 0.89\\ 
						   & $\#$epochs& 64.00 $\pm$ 31.91 & 55.67 $\pm$ 34.20\\
                           & $\sigma^2_{opt}$ & 1.98 $\pm$ 0.62 & 1.39 $\pm$ 0.81 \\\hline                         
\multirow{3}*{GMLVQ}    &  $\#$ prototype & 2.40 $\pm$ 0.62 & 2.70 $\pm$ 0.60\\ 
						   & $\#$epochs& 57.67 $\pm$ 32.87 & 31.67 $\pm$ 18.95\\
                           \hline                         
\end{tabular}
\end{table}

\begin{table}
\caption{Selected number of prototypes per class and $\sigma^2_{opt}$ on the ETH80 data set}
\label{tab:paraETH80}
\centering
\begin{tabular}{lllll}\hline
    % & \multicolumn{2}{c|}{GLVQ} & \multicolumn{2}{c|}{GRLVQ} &\multicolumn{2}{c|}{GMLVQ} &\multicolumn{2}{c|}{RGLVQ} \\\hline
No. of Categories & $\#$ prototype &  $\sigma^2_{opt}$ \\\hline
3 & 4.00 $\pm$ 1.30 &0.49 $\pm$ 0.02 \\
4 & 4.75 $\pm$ 1.21 &0.50 $\pm$ 0.01\\
5 & 4.95 $\pm$ 1.15 &0.50 $\pm$ 0.01\\
6 & 5.75 $\pm$ 0.44 &0.50 $\pm$ 0.00\\
7 & 5.75 $\pm$ 0.55 &0.50 $\pm$ 0.00\\
8& 5.85 $\pm$ 0.49 &0.5 $\pm$ 0.00 \\
\hline 
\end{tabular}
\end{table}

\begin{table}
\caption{Selected $\sigma^2_{opt}$, number of prototypes per class, and training epochs on motor imagery classification data sets}
\label{tab:paramotordata}
\centering
\begin{tabular}{llll}\hline
    % & \multicolumn{2}{c|}{GLVQ} & \multicolumn{2}{c|}{GRLVQ} &\multicolumn{2}{c|}{GMLVQ} &\multicolumn{2}{c|}{RGLVQ} \\\hline
Data set & $\#$ prototype & $\#$epochs&  $\sigma^2_{opt}$ \\\hline

S1 & 1 &20&3.5\\
S2 & 4 &20&1\\
S3 & 4 &20&1\\
S4 & 2 &100&8\\
S5 & 1 &20&4\\
S6 & 1 &20&3.5\\
S7 & 2 &50&5.5\\
S8 & 2 &20&2.5\\
S9 & 1 &100&5\\
\hline 
\end{tabular}
\end{table}

\end{appendix}

\newpage

\section*{Acknowledgements}

This work is supported by the National Natural Science Foundation of China (Grant No. 61803369, 51679213), the Natural Science Foundation of  Liaoning Provience of China (Grant No. 20180520025),the National Key Research and Development Program of China (Grant No. 2019YFC1408501),the Basic Public Welfare Research Plan of Zhejiang Province (LGF20E090004), and the  EC  Horizon  2020  ITN  SUNDIAL  (SUrvey  Network  for Deep  Imaging  Analysis  and  Learning),  Project  ID: 721463.

%Use unnumbered third level headings for the acknowledgements title.
%All acknowledgements go at the end of the paper.

%\subsubsection*{References}

%\bibliographystyle{plain}
\bibliographystyle{named}
\bibliography{reference}

\end{document}